\documentclass[]{bytedance_seed}

\usepackage[toc,page,header]{appendix}

\usepackage{minitoc}
\usepackage{amsfonts}
\usepackage[table]{xcolor}  
\usepackage{subcaption}
\usepackage{tikz}
\usetikzlibrary{fit,calc,tikzmark}

\usepackage{amsmath,amsfonts,bm,amssymb,mathtools,amsthm}
\usepackage{color,xcolor,placeins}
\usepackage{thm-restate}

\definecolor{orange}{HTML}{ff6c0c}
\definecolor{blue}{HTML}{1f77b4}
\definecolor{Gray}{gray}{0.85}
\definecolor{LightCyan}{rgb}{0.88,1,1}

\usepackage{url}
\usepackage{xkcdcolors}
\usepackage{mdframed}

\usepackage{microtype}
\usepackage{graphicx}
\usepackage{subcaption}
\usepackage{latexsym}
\usepackage{sidecap}
\usepackage{caption}
\usepackage{booktabs} 
\usepackage{amsfonts}       
\usepackage{nicefrac}       
\usepackage{comment}
\usepackage{enumitem}
\usepackage{wrapfig,tikz}
\usepackage{longtable}
\usepackage{bigstrut, tabularx, multirow, makecell, diagbox}
\usepackage[export]{adjustbox}
\usepackage{float}
\usepackage{stackrel}
\usepackage{color}
\usepackage{colortbl}
\usepackage{theoremref}
\usepackage{cases}
\usepackage{stmaryrd}
\usepackage{mathabx}
\usepackage{dsfont}
\usepackage{arydshln}
\usepackage{algorithm}
\usepackage{algpseudocode}

\makeatletter
\usepackage{xspace}
\def\@onedot{\ifx\@let@token.\else.\null\fi\xspace}
\DeclareRobustCommand\onedot{\futurelet\@let@token\@onedot}

\definecolor{blue1}{RGB}{0,128,255}
\definecolor{blue3}{RGB}{0,0,128}
\definecolor{darkpastelgreen}{rgb}{0.01, 0.75, 0.24}
\definecolor{cerulean}{rgb}{0.0, 0.48, 0.65}

\definecolor{darkgreen}{rgb}{0,0.6,0}

\def\eqref#1{equation~(\ref{#1})}

\def\1{\bm{1}}

\def\rvh{{\mathbf{h}}}

\def\rvm{{\mathbf{m}}}

\def\rvp{{\mathbf{p}}}
\def\rvq{{\mathbf{q}}}

\def\rvx{{\mathbf{x}}}

\def\rvy{{\mathbf{y}}}
\def\rvz{{\mathbf{z}}}

\def\vtheta{{\bm{\theta}}}

\def\vomega{{\bm{\omega}}}

\def\vphi{{\bm{\phi}}}
\def\vpsi{{\bm{\psi}}}

\DeclareMathAlphabet{\mathsfit}{\encodingdefault}{\sfdefault}{m}{sl}
\SetMathAlphabet{\mathsfit}{bold}{\encodingdefault}{\sfdefault}{bx}{n}

\def\gC{{\mathcal{C}}}

\def\gG{{\mathcal{G}}}

\def\sR{{\mathbb{R}}}

\newcommand{\ours}{EOSTok}
\definecolor{darkgreen}{RGB}{0,100,0}
\definecolor{darkred}{RGB}{139,0,0}

\definecolor{diffhl}{HTML}{EEEEEE}

\title{End-to-End Autoregressive Image Generation with 1D Semantic Tokenizer}
\author[1, 2, \S]{Wenda Chu}
\author[1, 2, \S]{Bingliang Zhang}
\author[1, 3, \S]{Jiaqi Han}
\author[1, \S]{Yizhuo Li}
\author[1]{Linjie Yang}
\author[2]{Yisong Yue}
\author[1]{Qiushan Guo}

\affiliation[1]{ByteDance Seed}
\affiliation[2]{California Institute of Technology}
\affiliation[3]{Stanford University}

\contribution[\S]{Work done at ByteDance Seed}

\abstract{
Autoregressive image modeling relies on visual tokenizers to compress images into compact latent representations. We design an end-to-end training pipeline that jointly optimizes reconstruction and generation, enabling direct supervision from generation results to the tokenizer. This contrasts with prior two-stage approaches that train tokenizers and generative models separately. We further investigate leveraging vision foundation models to improve 1D tokenizers for autoregressive modeling. Our autoregressive generative model achieves strong empirical results, including a state-of-the-art FID score of 1.48 without guidance on ImageNet 256×256 generation.
}

\date{\today}

\begin{document}
\maketitle

\resizebox{\textwidth}{!}{
\includegraphics{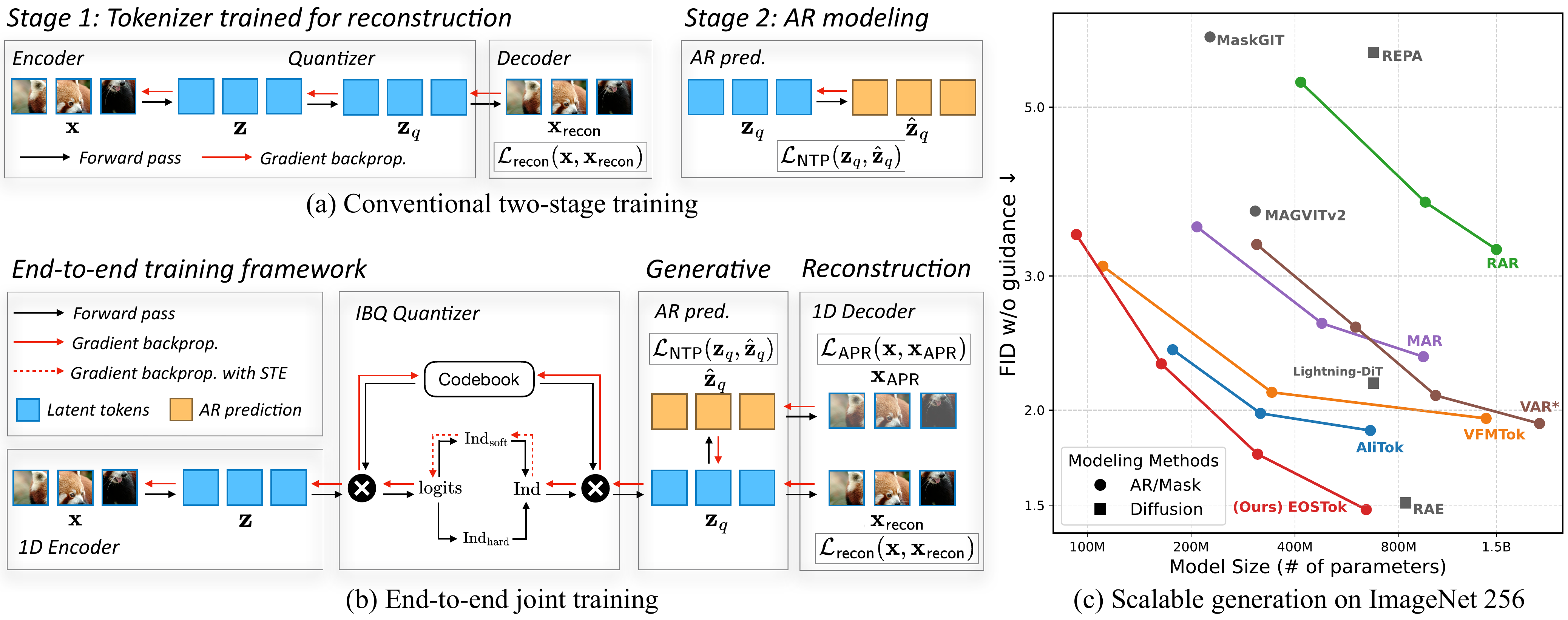}
}
\vspace{-15pt}
  \captionof{figure}{\textbf{Overview of \ours~that jointly trains an 1D vision tokenizer and an autoregressive generative model.} (a) Conventional paradigms employ a two-stage training strategy. The tokenizer is trained on a reconstruction task. An autoregressive generative model is then trained on the token sequences encoded by the frozen tokenizer. (b) Our training framework jointly optimizes the tokenizer and the AR model, enabling end-to-end generative supervision to guide the tokenizer. (c) Our~\ours~model enjoys a superior scaling ability on the ImageNet 1K 256$\times$256 generation benchmark. We compare~\ours~to generative models with AR, mask, or diffusion modeling. Numbers are collected without guidance except for those with (*).}
  
\vskip 0.15in

\section{Introduction}

Inspired by the remarkable success of autoregressive (AR) modeling in large language models~\cite{gpt3}, recent work has increasingly explored autoregressive visual generation via discrete visual tokens, positioning AR models as a compelling alternative to diffusion-based approaches~\cite{van2017neural,llamagen}.
However, most existing image generative models rely on 2D grid-structured tokenizers that preserve the spatial layout of pixel patches. Such 2D tokenizations induce inherently bidirectional dependencies among tokens, which are fundamentally misaligned with the unidirectional factorization required by raster-order autoregressive modeling.

Two broad classes of approaches have been explored to address this issue. One line of work retains 2D aligned tokenizers but designs new AR modeling schemes beyond raster-scan ordering, such as masked AR generation with random ordering~\cite{mar,maskgit} or next-scale prediction with multi-scale 2D tokenization~\cite{VAR}. In contrast, another line of work~\cite{titok,ge2023making,flextok,pcatok} seeks to build 1D tokenizers suitable for generative modeling. For example, TiTok~\cite{titok} concatenates image patches with  learnable query tokens, forcing these query tokens to compress global visual information into a compact 1D latent representation without imposing an explicit 2D spatial prior. This architecture was originally designed~\cite{titok} for the purpose of high compression rates (e.g., 32 tokens), which trades off reconstruction quality for the ease of generative modeling.
However, we argue that the removal of 2D structural dependency paves the way for designing 
visual tokens that naturally support vanilla autoregressive modeling, which does not necessarily rely on aggressive compressions.

To achieve this goal, we introduce an \textbf{\textit{end-to-end single-stage training paradigm}} that jointly optimizes for reconstruction and autoregressive generative modeling.
Unlike the conventional training paradigm that trains tokenizers only for reconstruction, we propose to jointly optimize reconstruction and generation, which enables direct generative feedback to the tokenizer. Moreover, we find that the next-token-prediction loss on \textit{discrete tokens} cannot determine the final generative quality in the \textit{pixel space}. To bridge this gap, we design an \textbf{\textit{Autoregressive Prediction Reconstruction}} (APR) loss that decodes the teacher-forcing prediction of the AR model into pixel space and computes a reconstruction loss. We show our end-to-end training pipeline improves the final generation quality and makes the latent space more autoregressive-predictable.

Furthermore, inspired by previous works~\cite{yao2025taming,yu2025repa} that use vision foundation models (VFM) to regulate the latent space of 2D tokenizers and improve diffusion models, we investigate how to effectively inject semantic VFM representations into a 1D tokenizer.
However, directly aligning the 1D sequential latent space to 2D VFM representations~\cite{yao2025taming} forces it to degenerate to a raster-ordered, patch-aligned sequence, leading to suboptimal performance. We thus propose an \textbf{\textit{implicit alignment}} strategy that aligns hidden patch embeddings to VFM representations instead. This strategy distills global semantic information from VFMs into the sequential latent space without enforcing spatial structure, resulting in substantially improved generative quality.

Putting it together, we present \ours, an \textbf{E}nd-to-end \textbf{O}ne-dimensional \textbf{S}emantic \textbf{Tok}enizer that jointly optimizes reconstruction, generation, and semantic alignment. Our method is effective and easily scalable, as the \ours-H model with 644M parameters achieves a state-of-the-art FID score of 1.48 without guidance on ImageNet-1K 256$\times$256 generation.

\section{Related Work}

\textbf{Image tokenizers.} 
Image tokenizers compress high-dimensional images into compact low-dimensional latent representations. Variational autoencoders~\cite{vae,rombach2022high} (VAEs) typically consist of an encoder that maps images $\rvx$ to latent embeddings $\rvz = \mathcal E(\rvx)$ and a decoder that reconstructs the image as $\hat\rvx = \mathcal D(\rvz)$, which are optimized using reconstruction loss and variation KL loss. While VAEs learn a continuous latent space, VQ-VAEs~\cite{van2017neural,razavi2019generating,esser2021taming} map images to discrete representations by adding a vector quantization module after the encoder, defining discrete latent codes as $\rvz = \mathcal Q(\mathcal E(\rvx))$. Later works improve VQ-VAE by applying residual quantization~\cite{rqvae} and dynamic quantization~\cite{dqvae}, and decreasing code dimension~\cite{llamagen,vitvqgan}. To scale the codebook size, recent approaches~\cite{magvitv2,FSQ,zhu2024scaling,IBQ} design new quantization algorithms to increase codebook utilization. Recently, 1D tokenizers~\cite{titok,ge2023making,flextok,pcatok} that encode 2D images into 1D sequences have gained more attention, which are further explained in the following paragraph.

\vspace{10pt}
\textbf{Autoregressive visual generation.} Early works on autoregressive modeling of images~\cite{rqvae,llamagen} predict image tokens in a raster-scan order, using tokenizers with latent spaces spatially aligned to 2D images. However, this strategy creates bidirectional dependencies among tokens, making it misaligned and thus suboptimal for causal autoregressive generation. To address this issue, MaskGIT~\cite{maskgit} and MAR~\cite{mar} propose masked AR modeling with bidirectional attention, while VAR~\cite{VAR} proposes autoregressive modeling for next-scale prediction.
Recently, TiTok~\cite{titok} and SEED~\cite{ge2023making} propose to extract 1D representations of images using learnable query tokens. The vision transformer~\cite{vit} (ViT) encoder appends these query tokens $\rvq$ to the 2D image patches $\rvx_{\text{patch}}$ and outputs only the query tokens $\rvz = \mathcal E_\vphi(\rvx_{\text{patch}} \oplus \rvq)$. To reconstruct images, the ViT decoder concatenates learnable mask tokens $\rvm_{\text{patch}}$ with the latent tokens, transforming the mask tokens back to images $\hat \rvx = \mathcal D_\vpsi( \rvm_{\text{patch}}\oplus \mathcal Q(\rvz))$. FlexTok~\cite{flextok} and Semanticist~\cite{pcatok} further introduce nested dropout on top of 1D latent tokens, enforcing important information to be represented by earlier tokens. 

\vspace{10pt}
\textbf{Representation from vision foundation models.} Leveraging semantic features from pre-trained vision foundation models (VFMs), e.g., DINO~\cite{oquab2023dinov2,simeoni2025dinov3} and CLIP~\cite{radford2021clip}, have achieved remarkable success in improving diffusion models. A line of work~\cite{yu2025repa,yao2025taming,leng2025repae} aligns the latent space (VA-VAE~\cite{yao2025taming}) or intermediate layers of diffusion models (REPA~\cite{yu2025repa}) to the corresponding VFM representations. Specifically, let $f$ be the pre-trained VFM and $\rvy = f(\rvx)$ be the VFM representation. The REPA loss is defined as 
$\mathcal L_{\mathsf{REPA}} = - \frac{1}{N} \sum_{n=1}^N\mathsf{sim} (h_\vomega(\rvh^{[n]}), \rvy^{[n]})),$
where $h_\vomega$ is a learnable MLP projector, $n$ is the patch index, and $\rvh$ is either the latent vector of the VAE~\cite{yao2025taming}, or the hidden state from an early layer of diffusion models~\cite{yu2025repa}. 
In contrast, another line of work~\cite{rae,svg,bi2025vision,chen2025aligning} directly substitutes the VAE encoder with a frozen, pre-trained vision encoder, optionally adding a lightweight learnable adaptation module.

\begin{figure*}[t]
    \centering
    \includegraphics[width=\textwidth]{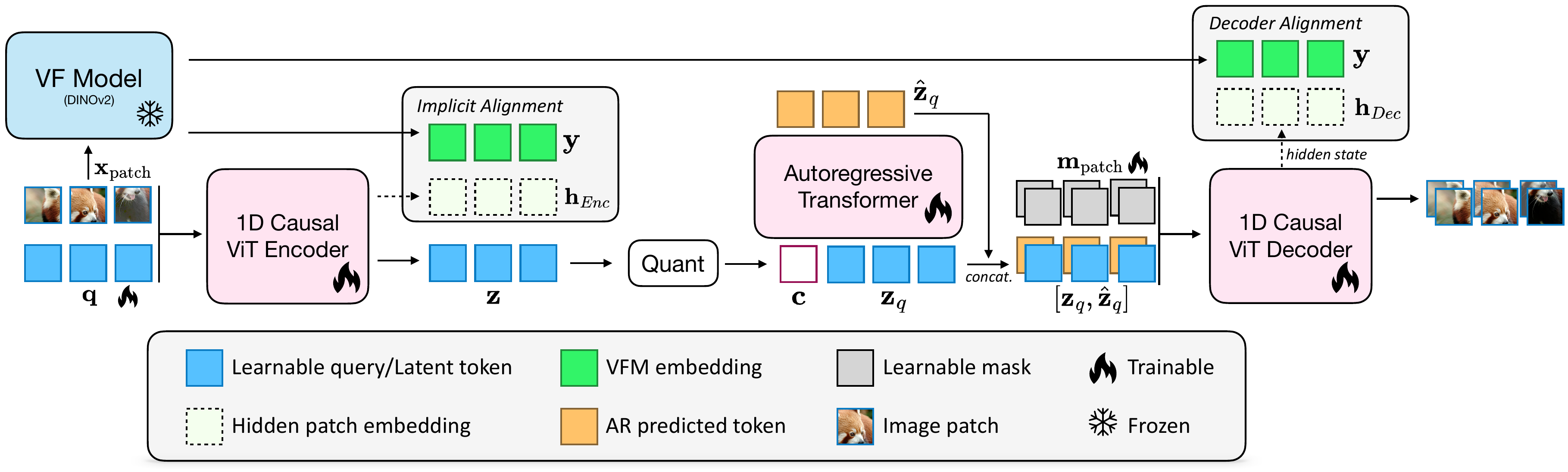}
    \caption{\textbf{Overall training pipeline of \ours.} The 1D ViT tokenizer and the autoregressive generative model are jointly trained. The training objectives include (1) \textbf{\textit{reconstruction loss}} $\mathcal L_{\mathsf{VQ-VAE}}$ that optimizes the tokenizer; (2) \textbf{\textit{generative loss}} (next token prediction) $\mathcal L_{\mathsf{NTP}}$ that optimizes AR model and supervises the latent space; (3) \textbf{\textit{AR prediction reconstruction}} (APR) loss (\cref{sec:joint-training}) $\mathcal L_{\mathsf{APR}}$ that decodes next token predictions to the pixel space and provides end-to-end generative feedback; and (4) \textbf{\textit{representation alignment loss}} $\mathcal L_{\mathsf{align}}$ (\cref{sec:1drepa}) that aligns both the latent space and decoder to semantic VFM embeddings. }
    \vspace{-10pt}
    \label{fig:method}
\end{figure*}

\section{Method}

Our goal is to produce a 1D vision tokenizer that compresses images to a sequential latent representation $\rvz$ that facilitates autoregressive modeling, which predicts $p(\rvz_n\mid \rvz_{<n})$. In this section, we first introduce our 1D ViT tokenizer architecture, and then introduce how end-to-end joint training (\cref{sec:joint-training}) and semantic VFM representation (\cref{sec:1drepa}) help us achieve this goal.

\subsection{1D Vision Transformer Tokenizer}
\label{sec:1dvit}

Our tokenizer is a ViT-based autoencoder with a discrete, 1D sequential latent space, which is designed following a similar structure as TiTok~\cite{titok}. As shown in~\cref{fig:method}, 2D-grid image patches $\rvx_{\text{patch}}\in \sR^{N\times D}$ are flattened and concatenated with $L$ learnable query tokens $\rvq\in \sR^{L\times D}$. This sequence is passed to a causal ViT encoder $\mathcal E_\vphi$, yielding $[\rvh_{\text{Enc}}, \rvz] = \mathcal E_{\vphi}([\rvx_{\text{patch}}, \rvq])$, where the hidden patch embedding $\rvh_{\text{Enc}}$ is discarded and only the 1D latent representation $\rvz$ is retained.
We employ vector quantization with IBQ~\cite{IBQ} and quantize the latent code to $\rvz_{q} = \mathcal Q(\rvz)$.
Symmetrically, the 1D decoder $\mathcal D_\vpsi$ is designed to take in $[\rvz_q, \rvm_{\text{patch}}]$ and reconstruct $[\varnothing, \rvx_{\text{recon}}]$, where $\rvm_{\text{patch}}\in \sR^{N\times D}$ are identical mask tokens.

\vspace{10pt}
\textbf{Quantizer.} The IBQ~\cite{IBQ} quantizer computes $\mathsf{logits} = [\rvz^T \gC_1, \ldots, \rvz^T \gC_K]\in \sR^K$ with $\gC\in \mathbb R^{K\times D}$ being the codebook. To enable gradient propagation, IBQ implements a straight-through estimation trick to compute code indices, i.e.,
\vspace{-5pt}
\begin{equation}
    \mathsf{Ind} = \text{onehot}(\arg\max \rvp) + [\rvp - \text{stopgrad}(\rvp)],
\end{equation}
where $\rvp = \mathrm{softmax}(\mathsf{logits})$. $\rvz_q = \mathsf{Ind}^T \gC$ is the quantized output representation. 

This 1D ViT architecture explicitly eliminates the 2D spatial prior from the tokenization process. 
Such decoupling allows us to design an image tokenizer that is inherently compatible with autoregressive generation. To this end, we propose an end-to-end training framework (\cref{sec:joint-training}) that jointly optimizes reconstruction and autoregressive generation. Furthermore, we investigate how incorporating semantic features from vision foundation models into 1D tokenizers can improve both reconstruction quality and generative performance (\cref{sec:1drepa}).

\subsection{Joint Training of Reconstruction and Generation}
\label{sec:joint-training}

Consider the task of training an image tokenizer with encoder $\mathcal E_\phi$ and decoder $\mathcal D_\psi$. The tokenizer is optimized by 
\begin{equation}
    \mathcal L_{\mathsf{VQVAE}}(\vphi, \vpsi) = \mathcal L_{\mathsf{recon}}(\rvx, \mathcal D_\vpsi(\rvz_q)) + \lambda_{\mathsf{reg}}\mathcal L_{\mathsf{reg}},
\end{equation}
where $\rvz_q = \mathcal Q(\rvz) = \mathcal Q(\mathcal E_\vphi(\mathcal \rvx))$ is the quantized latent representation of the image $\rvx$. $\mathcal L_{\mathsf{recon}}$ is a combination of $L_1/L_2$ loss, perceptual loss~\cite{lpips}, and GAN loss~\cite{esser2021taming}; while $\mathcal L_{\mathsf{reg}}$ regularizes the quantizer $\mathcal Q$, including commitment loss~\cite{lpips}, entropy loss~\cite{magvitv2}, and etc.

We study enhancing visual tokenizers for AR generation by jointly optimizing it for reconstruction and generation and enabling end-to-end generative supervision directly from AR modeling. The conventional paradigm trains a tokenizer for reconstruction in the first phase and a generative model in the second phase, with the tokenizer parameters frozen. However, we argue that freezing the tokenizer in the second phase prevents it from learning a representation better suited to the generative task.
This motivates us to design a \textit{\textbf{single-stage end-to-end}} training pipeline that jointly optimizes the tokenizer and the generative model from scratch. The overall loss function can be written as
\begin{equation}
    \mathcal L_{\mathsf{E2E}} (\vphi,\vpsi, \vtheta) = \mathcal L_{\mathsf{VQVAE}}(\vphi,\vpsi) + \lambda_{\mathsf{NTP}}\mathcal L_{\mathsf{NTP}}(\vphi, \vtheta),
\end{equation}
where $\vphi, \vpsi,\vtheta$ are parameters of the VAE encoder $\mathcal E_\vphi$, the decoder $\mathcal D_\vpsi$, and the AR generative model $\gG_{\vtheta}$, respectively. $\mathcal L_{\mathsf{NTP}}$ is the next token prediction loss, which depends on not only the AR model $\vtheta$ but the encoder $\vphi$ as well.

\definecolor{rowhl}{HTML}{E6E6FA}

\begin{figure}
    \centering
    \begin{minipage}{0.48\linewidth}
        \centering
        \includegraphics[width=\linewidth]{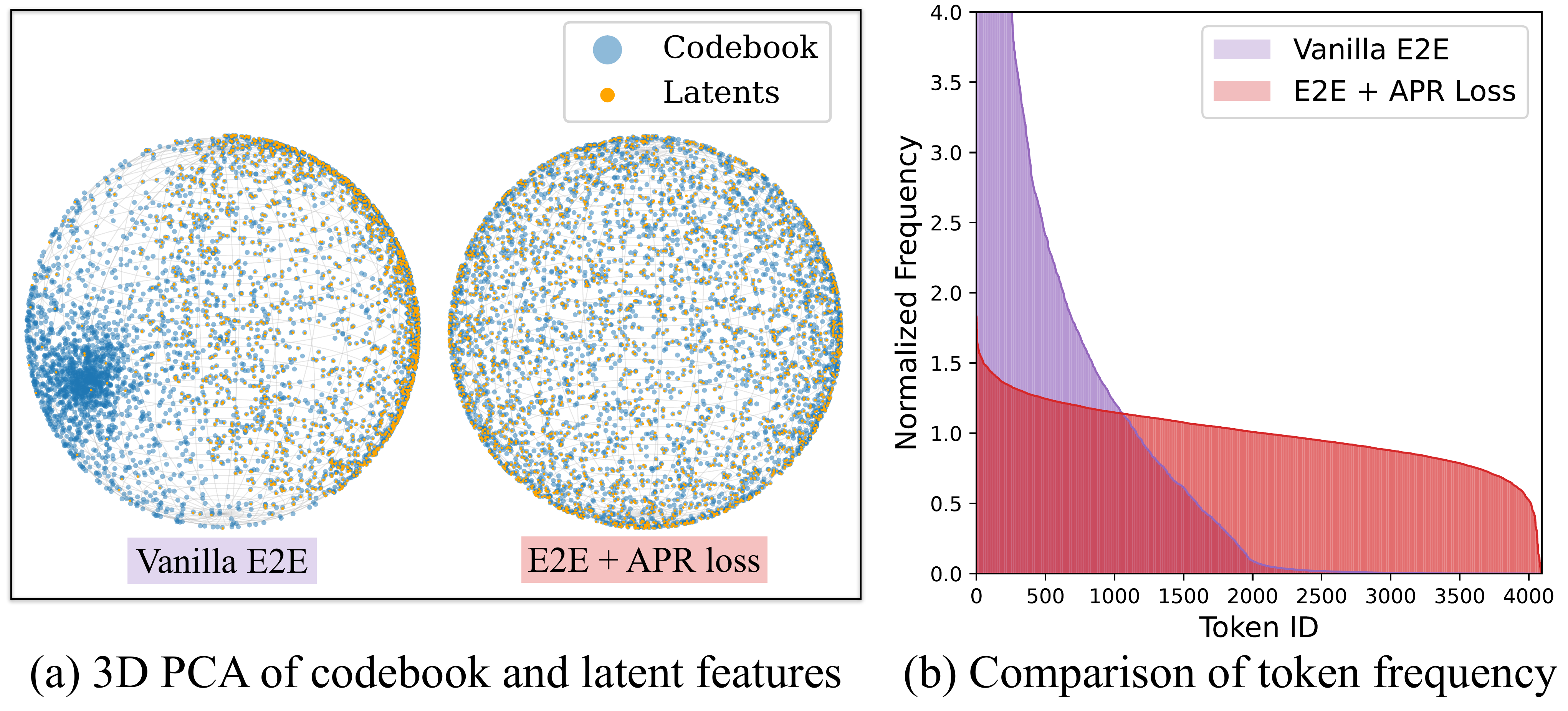}
        \captionof{figure}{\textbf{Next-token-prediction loss hacking the latent space.} (a) NTP loss encourages the tokenizer to use only a subset of tokens in the codebook, and causes the codebook to collapse. This is alleviated by the APR loss, which regulates the latent space by end-to-end feedback from the pixel space. (b) The tokenizer hacks the NTP loss by using very few tokens.}
        \label{fig:visual_e2e}
    \end{minipage}\hfill
    \begin{minipage}{0.48\linewidth}
        \centering
        \captionof{table}{\textbf{Joint training with APR loss prevents latent space collapsing and improves AR generation quality.} Numbers are reported by training a EOSTok-L model for 50 epochs. Images are generated \textit{without} classifier guidance and the prediction accuracy of the AR model is evaluated on the validation dataset. Code usage counts the code with a frequency of more than $5\%/K$ on the validation dataset.}
        \label{tab:ablate_e2e}
        \resizebox{\linewidth}{!}{%
        \begin{tabular}{lcccc}
        \toprule
             & rFID $\downarrow$ & gFID $\downarrow$ & AR Acc. & Code Usage \\
        \midrule
         Baseline & 1.09 & 3.82 & 11.8\% & 99.8\%\\
         Vanilla E2E & 4.92 & 8.01  & 30.2\% & 51.8\% \\
         \rowcolor{rowhl}
         + APR loss & \textbf{1.02} & \textbf{3.32} & 11.9\% & 99.7\% \\
         \bottomrule
        \end{tabular}}
    \end{minipage}
\end{figure}

\vspace{10pt}
\textbf{Gradient propagation.} During joint training, the AR model is trained on quantized tokens that are not naturally differentiable with respect to the tokenizer. To enable gradient supervision from AR modeling, we modify the embedding layer of our AR model so that it takes in probability $\mathsf{Ind}\in \mathbb R^{L\times K}$ and compute the embedding as $\rvh = \mathsf{Ind}^T \mathsf{Embed}$ instead of a look-up operation. This enables gradient backward of $\mathcal L_{\mathsf{NTP}}$ to the VAE encoder and the codebook.

\vspace{10pt}
\textbf{Gap between NTP loss and generation quality.} The biggest challenge of jointly training tokenizer and AR model is that the next-token-prediction loss is not an end-to-end objective. The NTP objective is defined on discrete token space that is constantly changing during training, and it cannot reflect the final generation quality in the pixel space. 

To demonstrate this gap, we train a model using the joint training framework, which we refer to as \textit{Vanilla E2E}. As shown in~\cref{tab:ablate_e2e}, supervising the tokenizer with $\mathcal L_{\mathsf{NTP}}$ significantly boost the AR prediction accuracy (from 10.8\% to 30.2\%). However, the NTP loss hacks the tokenizer by \textit{collapsing the latent space into using very few tokens}, resulting in a dramatic decrease in codebook usage, rFID, and gFID.

We further visualize its codebook by a principal component analysis and projecting it onto a 3D sphere in~\cref{fig:visual_e2e}. As shown in~\cref{fig:visual_e2e}a, the codebook distribute unevenly in the latent space, and the latent embedding of images could only match a small fraction of codes. This is corroborated by~\cref{fig:visual_e2e}b, where the frequency distribution is highly skewed towards a small subset of tokens.

\vspace{10pt}
\textbf{Bridging NTP and generative quality.} To solve this issue, we propose a simple yet effective \textbf{\textit{autoregressive prediction reconstruction}} \textbf{\textit{(APR)}} loss to bridge the gap between NTP loss and the overall generation quality. We take the predicted tokens of AR model in teacher forcing, decode them directly to pixels using the decoder, and match with the ground-truth image. The loss function can be written as
\begin{equation}
    \mathcal L_{\mathsf{APR}}(\vphi,\vpsi,\vtheta) = \|\rvx - \mathcal D_\vpsi(\mathcal G_\vtheta(\rvz_q))\|_2^2,
\end{equation}
where $\rvz_q = \mathcal Q(\mathcal E_{\vphi}(\rvx))$ are quantized tokens. Similar to $\mathcal L_{\mathsf{recon}}$, this MSE objective can be enhanced with perceptual loss, e.g., LPIPS~\cite{lpips}. During training, we concatenate the AR prediction $\hat \rvz_q = \mathcal G_\vtheta(\rvz_q)$ with $\rvz_q$ along the batch dimension, and pass them together to the decoder.

The APR loss provides end-to-end generative supervision to the tokenizer directly from the pixel space, and regulates next token prediction loss to be meaningful.

This is confirmed by~\cref{fig:visual_e2e}a, where end-to-end training with APR loss resolves the latent collapse issue caused by back-propagating the NTP loss only. 
As shown in~\cref{tab:ablate_e2e}, applying APR loss to the end-to-end training framework  improves the overall generation quality, notably lowering the gFID score from 8.01 to 3.32.

\subsection{Introduce Semantic Representation to Tokenizers}
\label{sec:1drepa}

\begin{figure*}[t]
    \centering
    \includegraphics[width=\linewidth]{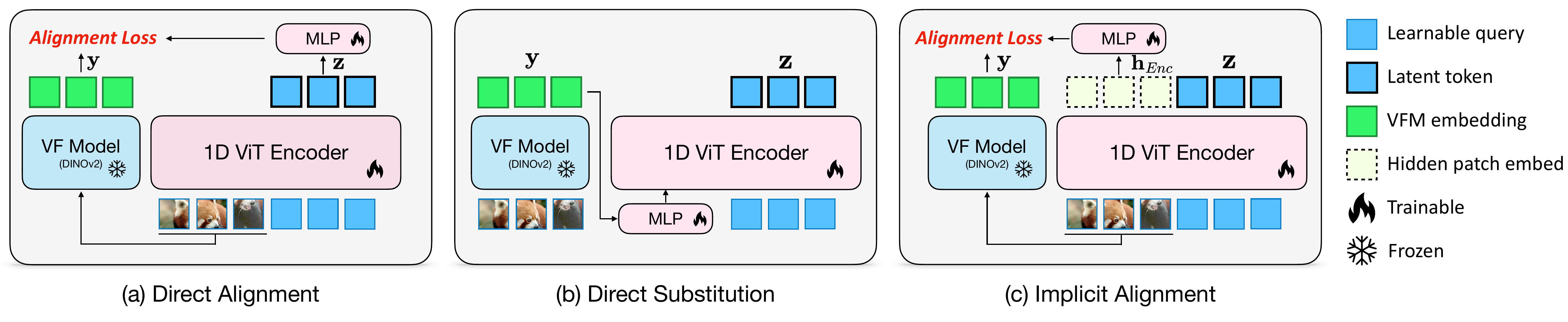}
    \caption{\textbf{Variants of representation injection into 1D VAE encoder.} Image patches are passed to a vision foundation model (e.g., DINOv2) to extract semantic VFM embeddings. To inject this representation into the 1D ViT encoder, (a) \textbf{\textit{direct Alignment}} means aligning the latent space of tokenizers to the VFM embedding; (b) \textbf{\textit{direct substitution}} replaces 2D image patches with the VFM embedding; and (c) \textbf{\textit{implicit alignment}} aligns the hidden patch embedding to VFM representation.}
    \label{fig:1drepa}
\end{figure*}

Inspired by the success of representation alignment~\cite{yu2025repa,leng2025repae,yao2025taming} in diffusion models,
we explore the idea of using semantic features from a vision foundation model $f$ to help improve visual tokenization. 
To this end, we comprehensively investigate several variants of injecting the representation $f(\rvx)$ into both encoder and decoder of the 1D tokenizer.

\vspace{10pt}
\textbf{Types of representation injection into encoders.} 
We first study three methods for injecting semantic representation into the 1D ViT encoder, as illustrated in~\cref{fig:1drepa}. 

\begin{itemize}[leftmargin=10pt,topsep=0pt,itemsep=5pt]
    \item \textbf{\textit{Direct alignment.}} Latent tokens $\rvz$ are directly aligned to pre-trained representation $\rvy = f(\rvx)$, similar to VA-VAE~\cite{yao2025taming}. Its loss function is defined as 
    \vspace{-5pt}
    \begin{equation}
        \mathcal L_{\mathsf{direct}}(\vomega,\vphi) = -\frac{1}{L}\sum_{\ell=1}^L\mathsf{sim}(h_\vomega(\rvz^{[\ell]}), \mathcal I(\rvy)^{[\ell]}),
    \end{equation} 
    where $\ell$ is the token index and $\mathcal I$ interpolates the 2D features from $\sR^{N\times D}$ to $\sR^{L\times D}$, and $\mathsf{sim}$ measures cosine similarity.
    This loss enforces the 1D latent codes $\rvz$ to match the spatially aligned features $f(\rvx)$, which inevitably leaks the 2D spatial prior to the 1D tokenizer.
    \item \textbf{\textit{Direct substitution.}} An alternative way to leverage visual representation of VFMs is to use them directly as an encoder, which has been explored in training diffusion models~\cite{rae,svg,bi2025vision,chen2025aligning}. As demonstrated in~\cref{fig:1drepa}(b), we replace image patches $\rvx_{\text{patch}}$ by the projected VFM features, i.e., $\mathsf{MLP}(f(\rvx))$, concatenate them with learnable quries $\rvq$, and pass them to the 1D ViT encoder. 
    \item \textbf{\textit{Implicit alignment.}} Instead of directly aligning latent tokens, we consider aligning the hidden patch embeddings $\rvh_\text{Enc}$ to the VFM representations $f(\rvx)$. The 1D latent codes $\rvz$ are not forced to match the VFM representation, but can still extract semantic information from the aligned 2D hidden embedding. We name this scheme implicit alignment, whose loss function is defined as 
    \begin{equation}
        \mathcal L_{\mathsf{implicit}} (\vomega, \vphi) = -\frac{1}{N}\sum_{n=1}^N \mathsf{sim}(h_\vomega(\rvh_{\text{Enc}}^{[n]}), \rvy^{[n]}). 
    \end{equation}
\end{itemize}

\vspace{10pt}
\textbf{\textit{Decoder alignment.}} In addition to injecting semantic representation into encoders, we further examine strategies of aligning the decoder to the VFM representation.  
We hypothesize that the reconstruction task of an 1D ViT decoder is much harder than a 2D ViT decoder, as it requires to recover pixels 
whose information distributes globally, instead of locally aligns with, the latent token sequence. This is more similar to a conditional generation task, rather than a reconstruction task. Therefore, applying representation alignment to 1D ViT decoders could help its convergence, as it does to accelerate diffusion models training. 
We thus take inspiration from~\citet{yu2025repa} and extract the hidden features of the mask tokens from the $k$-th layer of the decoder, denoted as $\rvh_{\text{Dec}}$, and align it with the VFM features $\rvy = f(\rvx)$.

\begin{wraptable}{r}{0.5\linewidth}  
    \centering
    \vspace{-\baselineskip}           
    \caption{\textbf{Quantitative comparison of injecting representation to 1D and 2D ViT tokenizers.} Numbers are reported by training a \ours-L model for 50 epochs. We use 256 query tokens for 1D tokenizer and to match the sequence length of 2D tokenizer for a fair comparison. Images are generated \textit{without} classifier guidance and the prediction accuracy of the AR model is evaluated on the validation dataset. Baseline is trained without VFM representations.}
    \label{tab:1drepa}
    \resizebox{\linewidth}{!}{%
    \begin{tabular}{lccc}
    \toprule
     & rFID $\downarrow$ & gFID $\downarrow$ & AR Acc. $\uparrow$\\
    \midrule
    \textbf{\textit{1D Tokenization}} & & &\\
    Baseline & 1.75 & 12.27 & 7.8\%\\
    + Decoder alignment & 1.12 & 5.68 & 8.2\% \\
    + (a) Direct alignment & \textbf{0.98} & 5.98 & 8.5\%\\
    \hspace{5.2pt} (b) Direct substitution & 1.05 & 4.89 & \textbf{12.1\%} \\
    \rowcolor{rowhl}
    \hspace{5.2pt} (c) Implicit alignment & 1.02 & \textbf{3.32} & 11.9\% \\
    \midrule
    \textbf{\textit{2D Tokenization}} \textit{for reference} & & & \\
    Baseline & 1.52 & 12.51 &  5.2\% \\
    + Decoder \& direct alignment & 0.87 & 6.06 & 7.9\% \\
    \bottomrule
    \end{tabular}}
\end{wraptable}

\vspace{10pt}
\textbf{Semantic alignments improve AR generative quality.}
We test these approaches of injecting VFM representations on ImageNet generation, comparing them to the baseline without using VFMs. As shown in~\cref{tab:1drepa}, applying \textbf{\textit{decoder alignment}} significantly improves the 1D tokenizer on reconstruction, reflecting on both rFID and gFID. However, the AR prediction accuracy increases marginally, so it does not make AR modeling easier. 

For leveraging VFM on 1D ViT encoder, all three approaches slightly improve the reconstruction quality over the baseline. 
However, applying \textbf{\textit{direct alignment}} worsens the generative quality, which supports our hypothesis that enforcing 2D spatial structures on the latent space is detrimental to the AR modeling. 
Approaches (b) and (c) abstain from using a 2D spatial prior during training, among which \textbf{\textit{implicit alignment}} significantly improves generation quality. Moreover, applying both approaches improves the prediction accuracy of the AR generative model by a large margin, indicating that a more AR-generation-friendly tokenization can be learned through semantic guidance during training.

\section{Experiments}
\begin{figure*}[t]
  \centering
  \begin{subfigure}[t]{0.32\linewidth}
    \centering
    \includegraphics[width=\linewidth]{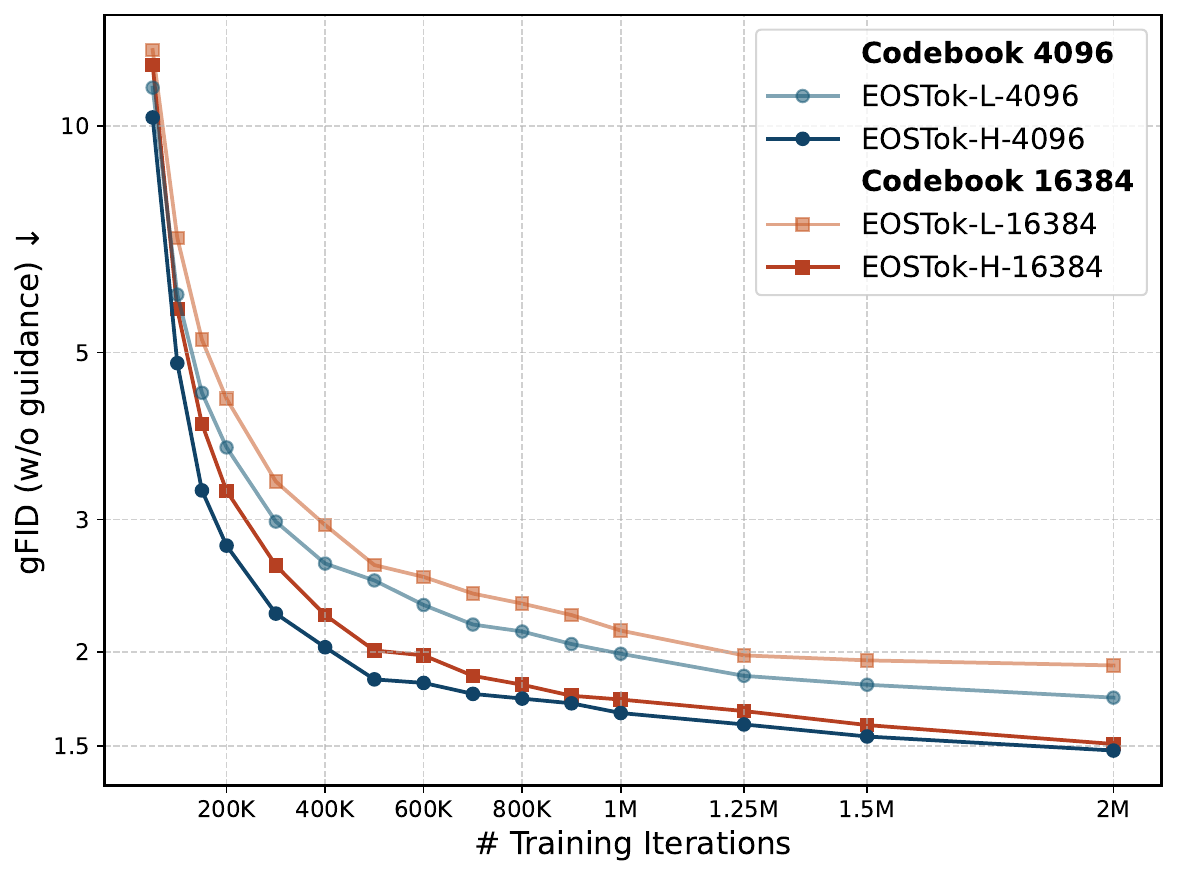}
    \caption{gFID vs. training iterations.}
    \label{fig:gfid_converge}
  \end{subfigure}
  \hfill
  \begin{subfigure}[t]{0.32\linewidth}
    \centering
    \includegraphics[width=\linewidth]{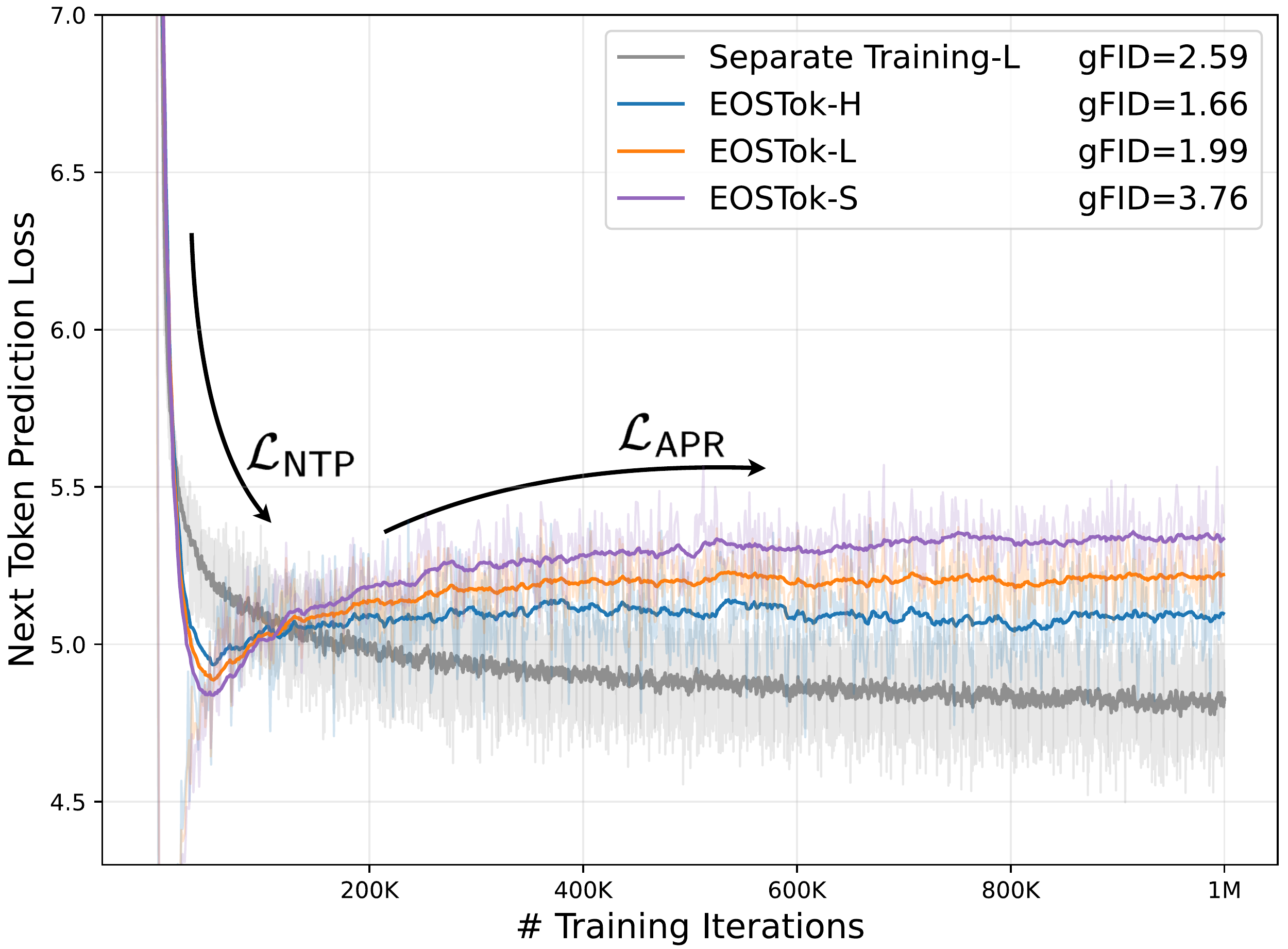}
    \caption{Next token prediction loss $\mathcal L_{\mathsf{NTP}}$}
    \label{fig:ce_loss}
  \end{subfigure}
  \hfill
  \begin{subfigure}[t]{0.32\linewidth}
    \centering
    \includegraphics[width=\linewidth]{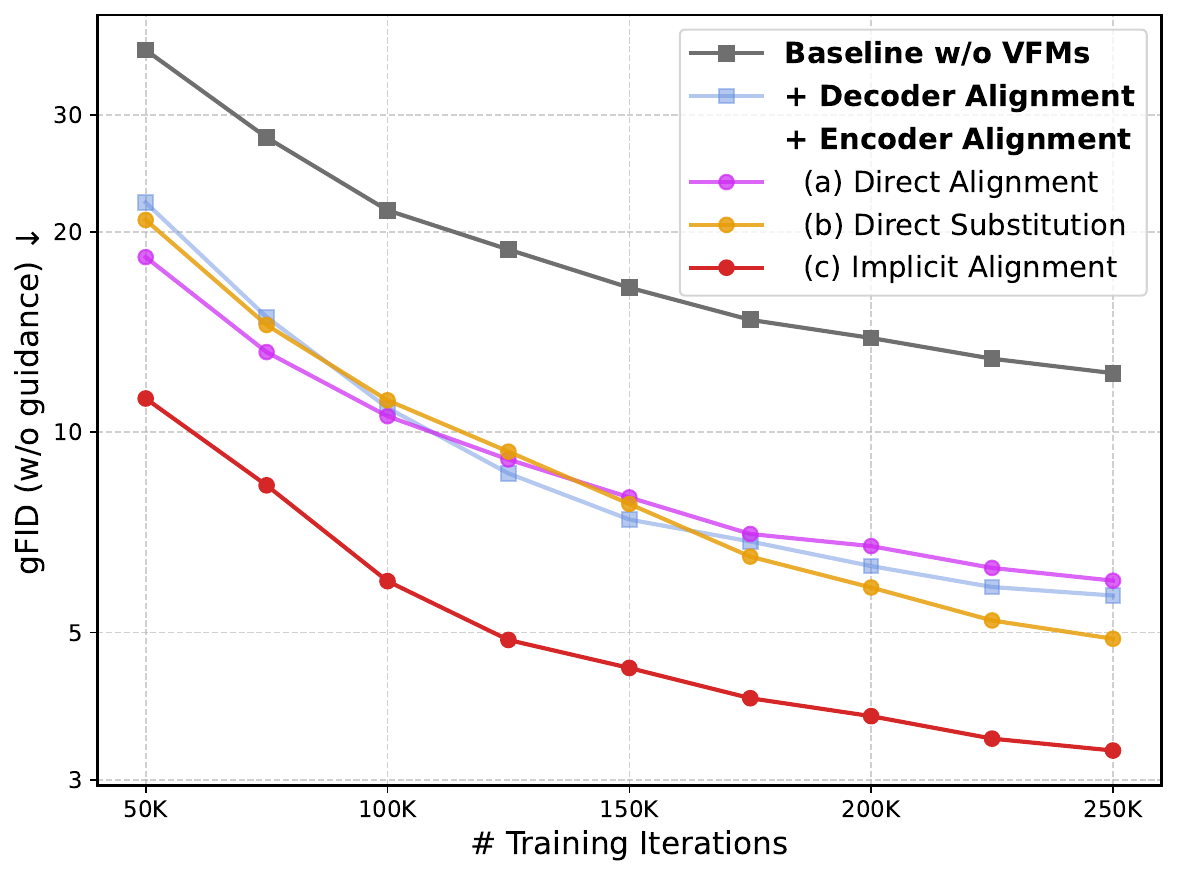}
    \caption{The gFID curves of different semantic representation injections.}
    \label{fig:ph}
  \end{subfigure}
  \caption{\textbf{Learning curves of \ours.} (a) We plot the generative FID score evaluated during training. For both codebook sizes $K$ of $4096$ and $16384$, scaling up the \ours~model consistently improves the overall generation quality. Moreover, using a larger model size (e.g., \ours-H) closes the gap between $K=4096$ and $K=16384$. (b) We plot the loss curves for $\mathcal L_{\mathsf{NTP}}$ across model sizes (S, L, and H). We compare our end-to-end training with two-stage separate training on model size L. Although separate training achieves lower NTP training loss, our framework achieves better overall generation quality by enabling end-to-end feedback, i.e., $\mathcal L_{\mathsf{APR}}$. (c) The convergence of our model using different semantic alignment strategies. Among the three strategies considered for injecting semantic representation into the encoder, \textit{implicit alignment} is the most effective.}
  \label{fig:training_curve}
\end{figure*}
\definecolor{rowhl}{HTML}{E6E6FA}

\begin{table*}[t]
\centering
\caption{\textbf{ImageNet 256 Generation Results.} We categorize existing generative models based on the visual tokenizer used, including \textbf{\textit{2D continuous}} tokenizers with latent diffusion and masked AR models, \textbf{\textit{2D discrete}} tokenizers with AR models and mask models, and \textbf{\textit{1D}} tokenizers with AR models. \ours~achieves the state-of-the-art performance on generation without using classifier guidance.}

\label{tab:main_result}
\resizebox{\textwidth}{!}{%
\begin{tabular}{l@{\hspace{5pt}}l@{}rlcl@{}rcccc}
\toprule
\multirow{2}{*}{\textbf{Method}} &
\multicolumn{4}{c}{\textbf{Tokenizer}} &
\multicolumn{2}{c}{\textbf{Generator}} &
\multicolumn{2}{c}{\textbf{w/o guidance}} &
\multicolumn{2}{c}{\textbf{w/ guidance}}  \\
\cmidrule(lr){2-5}\cmidrule(lr){6-7}\cmidrule(lr){8-9} \cmidrule(lr){10-11}
& \textbf{Type} & \#\textbf{Params} & \#\textbf{Tokens} & \textbf{rFID} $\downarrow$ &  \textbf{Type}  & \#\textbf{Params} & 
\textbf{gFID$\downarrow$} &  \textbf{IS$\uparrow$} & 
\textbf{gFID$\downarrow$} &  \textbf{IS$\uparrow$} \\
\midrule
\midrule
\rowcolor{rowhl}
\multicolumn{11}{c}{\textit{\textbf{2D Continuous Latent Space}}} \\
\midrule\midrule
LDM-4~\cite{rombach2022high} & SD-VAE & 55M & 64$\times$64 & 0.27 & Diff. & 400M & 10.56 & 103.5 & 3.60 & 247.7 \\
DiT-XL/2~\cite{peebles2023scalable} & SD-VAE & 84M & 32$\times$32 & 0.62 & Diff. & 675M & 9.62 & 121.5 & 2.27 & 278.2 \\
REPA-XL/2~\cite{yu2025repa} & SD-VAE & 84M & 32$\times$32 & 0.62 & Diff. & 675M & 5.90 & 157.8 & 1.42 & 305.7 \\
Lightning-DiT-XL~\cite{yao2025taming} & VA-VAE & 84M & 32$\times$32 & 0.28 & Diff.  & 675M & 2.17 & 205.6 & 1.35 & 295.3 \\   
MAR-L~\cite{mar} & SD-VAE & 66M & 16$\times$16 & 0.87 & MAR Diff. & 479M & 2.60 & 221.4 & 1.78 & 296.0 \\
\midrule\midrule
\rowcolor{rowhl}
\multicolumn{11}{c}{\textit{\textbf{2D Discrete Tokenization}}}\\
\midrule\midrule
VQGAN~\cite{esser2021taming} & VQ & 23M & 16$\times$16 & 4.98 & AR & 1.4B & 15.78 & 74.3 & - & - \\
RQTrans~\cite{rqvae} & RQ & 66M  & 8$\times$8$\times$4 & 3.20 & AR & 1.4B & 8.71 & 119.0 & 3.89 & 337.5  \\
DQTrans~\cite{dqvae} & DQ & 48M & 640 & 4.08 & AR&  655M & 5.11 & 178.2 & - & - \\
MaskGIT~\cite{maskgit} & VQ & 66M & 16$\times$16 & 2.28 & Mask & 227M & 6.18 & 182.1 & - & -\\
MAGVIT-v2~\cite{magvitv2} & LFQ & 133M & 16$\times$16 & 1.17 & Mask & 307M & 3.65 & 200.5 & 1.78 & 319.4 \\
LlamaGen-XL~\cite{llamagen} & VQ & 72M & 16$\times$16 & 0.94 & AR & 775M & 14.77 & 80.8 & 2.62 & 244.1\\
RAR-L~\cite{yu2025randomized} & VQ & 66M & 16$\times$16 & 2.28 & AR & 461M  & 5.39 & 149.1 & 1.70 & 299.5 \\
IBQ-L~\cite{IBQ} & IBQ & 128M & 16$\times$16 & 1.37 & AR & 649M & - & - & 2.45 & 267.5 \\ 
VAR-d20~\cite{VAR} & MSRQ & 109M & 680 & 0.90 & VAR & 310M & - & - & 2.57 & 302.6  \\
AliTok-L~\cite{wu2025alitok} & VQ & - & 16+16$\times$16 & 0.86 & AR & 318M & 1.98 & 200.8 & 1.38 & 326.2 \\
\midrule\midrule
\rowcolor{rowhl}
\multicolumn{11}{c}{\textit{\textbf{1D Tokenization}}}\\
\midrule\midrule
TiTok-L-32~\cite{titok} & 1D VQ & 641M & 32 & 2.21 & Mask & 177M & 3.15 & 173.0 & 2.77 & 199.8 \\
FlexTok d18-18~\cite{flextok} & 1D FSQ Flow & 950M & 1-256 & 1.61 & AR & 1.33B & - & - & 2.02 & - \\
Semanticist~\cite{pcatok} & 1D VAE Flow & - & 1-256 & 0.72 & AR Diff. & 343M  & - & - & 2.57 & 254.0 \\
GigaTok~\cite{gigatok} & 1D VQ & 622M & 256 & 0.81 & AR & 111M & - & - & 3.26 & 221.0 \\  
SpectualAR-d20~\cite{huang2025spectralar} & 1D VQ & - & 64 & 4.03 & AR & 600M & - & - & 2.49 & 305.4 \\ 
VFMTok~\cite{zheng2025vision} & 1D VQ & - & 256 & 0.89 & AR & 343M & 2.11 & 230.1 & 2.75 & 278.8 \\
ResTok~\cite{zhang2026restok} & 1D VQ & 662M & 128 & 1.28& HAR & 326M & - & - & 2.34 & 257.8\\[3pt]

\textbf{\ours-S (Ours)} & 1D IBQ & 165M & 256 & 0.74 & AR & 93M & 3.50 & 155.7 &  2.57 & 211.5 \\ 
\textbf{\ours-B (Ours)} & 1D IBQ & 165M & 256 & 0.73 & AR & 164M &  2.38 & 185.6 & 1.98 & 220.1 \\ 
\textbf{\ours-L (Ours)} & 1D IBQ & 165M & 256 & 0.73 & AR & 312M & 1.74 & 210.2  & \textbf{1.35} & 236.5\\  
\textbf{\ours-H (Ours)} & 1D IBQ & 388M & 256 & \textbf{0.71} & AR & 644M & \textbf{1.48} & \textbf{239.5} & 1.38 & 265.7\\
\bottomrule
\end{tabular}
}
\end{table*}

\subsection{Experimental Setup}

\textbf{Model Architecture.} As described in~\cref{sec:1dvit}, we use a 1D ViT tokenizer with similar architecture as~\citet{titok}, and adopt IBQ~\cite{IBQ} as the vector quantization module. Unless specified, all models are trained with a $K=4096$ codebook size, a $L=256$ token sequence length, and a $d=64$ latent dimension. We design the attention maps of both the encoder and decoder to be bidirectional over patch tokens but causal over learnable queries. 
To stabilize training, we apply $\ell_2$-normalization to both the latent vectors before quantization and the IBQ codebook, projecting them to the $K$-dimensional unit sphere. 
We build our AR generative model based on LlamaGen~\cite{llamagen}, with an additional shared global AdaLN modulation with per-block learnable biases~\cite{chen2023pixartalpha}.
We use DINOv2-ViT-L~\cite{oquab2023dinov2} for representation alignment.

\vspace{10pt}
\textbf{Training.} We train our model on ImageNet-1K 256$\times$256~\cite{deng2009imagenet} for 400 epochs using Adam optimizer~\cite{kingma2014adam} with a batch size of 256 and BF16 precision. We train our model on four different sizes (S, B, L, and H), while jointly scaling the tokenizer and the AR generative model. Detailed configurations are provided in \cref{appendix:model_config}. We use L2, LPIPS~\cite{lpips}, and GAN loss~\cite{gan,sauer2023stylegan} for reconstruction $\mathcal L_{\mathsf{recon}}$, and use L2 and LPIPS for APR loss $\mathcal L_{\mathsf{APR}}$. We stabilize the discriminator of GANs by LeCam divergence~\cite{tseng2021regularizing}.

\vspace{10pt}
\textbf{Evaluation.} We use the evaluation code from~\citet{dhariwal2021diffusion} and report  Fr\'echet Inception Distance~\cite{heusel2017gans} (both rFID and gFID), and Inception Score (IS)~\cite{salimans2016improved} to evaluate the reconstruction and generation quality. We adopt AutoGuidance~\cite{karras2024guiding} in replace of classifier-free guidance~\cite{ho2022classifier}.

\subsection{Generation Results on ImageNet 256$\times$256}

\begin{figure}[H]
    \centering
    \begin{minipage}[t]{0.4\linewidth}
    \vspace{0pt}
        \centering
        \includegraphics[width=\linewidth]{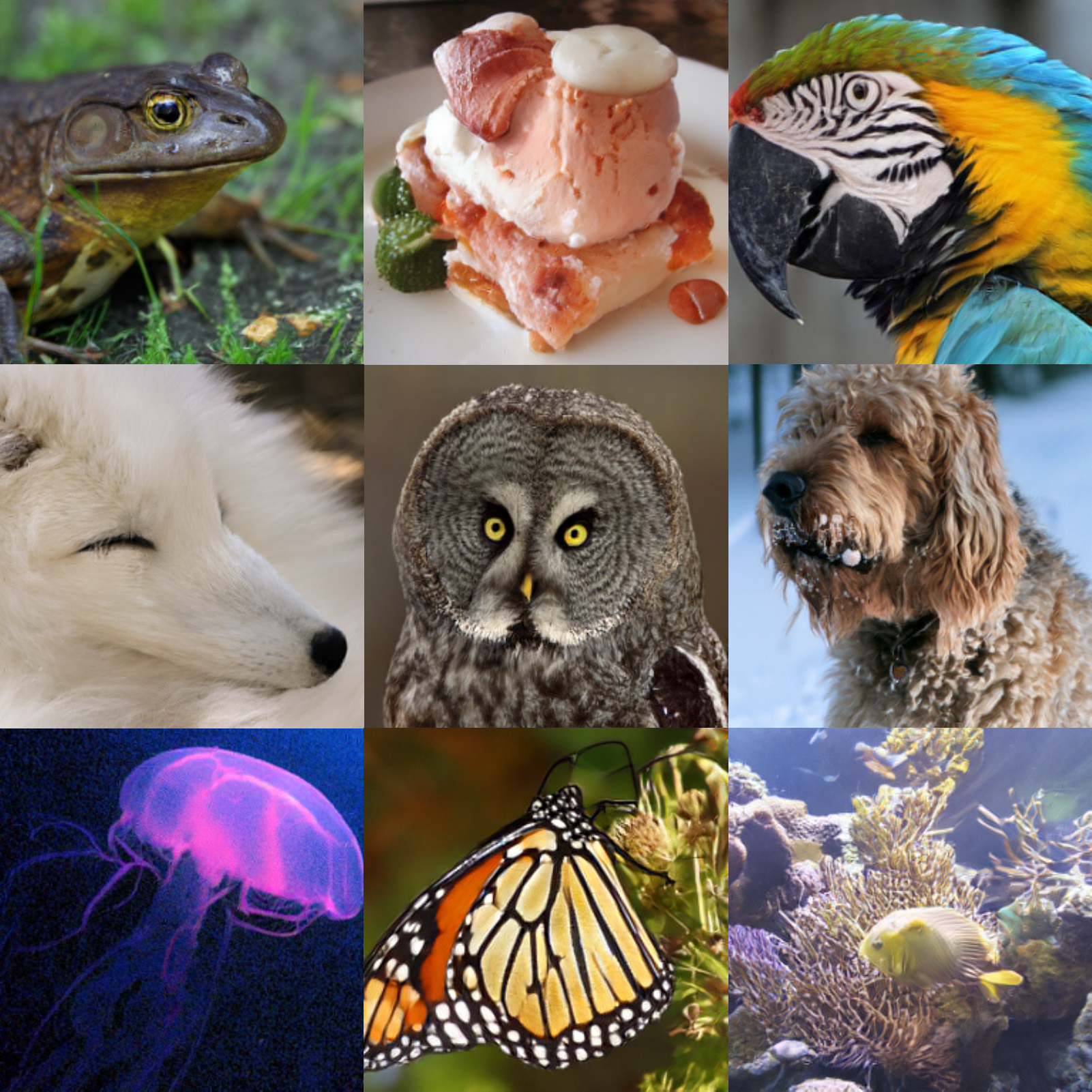}
        \caption{\textbf{Qualitative results on ImageNet 256 generation.}}
        \label{fig:visualize}
    \end{minipage}\hfill
    \begin{minipage}[t]{0.52\linewidth}
    \vspace{0pt}
        \centering
        \captionof{table}{\textbf{Training AR models on token sequences with different ordering.} Token sequences are from a tokenizer that is pre-trained using our end-to-end training framework.}
        \label{tab:ablate_order}
        \resizebox{\linewidth}{!}{%
        \begin{tabular}{rccc}
        \toprule
            & Original & Reversed & Random  \\
            \midrule
            gFID (w/o guidance) $\downarrow$ & \textbf{4.10} & 10.27 & 7.81 \\
            AR Accuracy $\uparrow$ & \textbf{10.3\%} & 9.5\% & 9.8\%\\
            \bottomrule
        \end{tabular}}

        \vspace{1em}   

        \captionof{table}{\textbf{Ablation on the choice of sequence length.} Increasing sequence length consistently improves the model's ability for reconstruction but could hurt generation quality if the sequence goes too long.}
        \label{tab:ablate_seq}
        \resizebox{\linewidth}{!}{%
        \begin{tabular}{lccccc}
        \toprule
            Token sequence length & 32 & 64 & 128 & 192 & 256 \\
            \midrule
            \hspace{1.4pt}rFID $\downarrow$ & 17.50 & 1.94 & 1.32 & 1.08 & \textbf{1.02} \\
            gFID (w/o guidance) $\downarrow$ & 22.37 & 3.18 & 3.09 & \textbf{3.04} & 3.32 \\
            \bottomrule
        \end{tabular}}
    \end{minipage}
\end{figure}

We demonstrate the effectiveness of our end-to-end training framework and compare \ours~with state-of-the-art methods on ImageNet-1K 256$\times$256 benchmark. As shown in~\cref{tab:main_result}, our \ours-L model achieves a 1.74 gFID \textit{without} guidance using only 312M parameters, outperforming the results of competitive 1D tokenization baselines even with classifier guidance. Moreover, our \ours~tokenizer has a much better reconstruction quality in terms of rFID, as the \ours-L tokenizer achieves an rFID of 0.73 with 165M parameters. This is especially credited to the use of semantic VFM representation alignment on the 1D ViT decoder, which helps the convergence of 1D tokenizer, as shown by the ablation study in~\cref{tab:1drepa}. We also provide some representative images generated by \ours~in~\cref{fig:visualize}.

\textbf{Scaling behavior.} We train~\ours~ with four different model sizes, from 93M to 644M. As shown in~\cref{fig:gfid_converge}, scaling the AR generative model consistently improves the generation quality as measured by FID. In \cref{fig:ce_loss}, we plot the learning curve of the cross-entropy loss, $\mathcal L_{\mathsf{NTP}}$, where scaling up the AR model lowers the converging point of the CE loss.
Moreover, we jointly scale the ViT tokenizer and the AR model \ours~to 388M and 644M parameters, respectively, for a total of 1B trainable parameters, enabling the \ours-H model to achieve a state-of-the-art gFID of 1.48 without guidance.

\subsection{Impact of End-to-end Training}

\textbf{Facilitating AR modeling.} To test the hypothesis that end-to-end training with APR loss builds a more AR generation-friendly latent space, we design an experiment as follows. We take an \ours-L tokenizer trained for 50 epochs, freeze it, and then train another AR model from scratch to generate the corresponding discrete visual tokens. We consider three different setups for the AR model:
\begin{itemize}[leftmargin=30pt,topsep=5pt,itemsep=5pt]
    \item [(a)]training on original token sequences;
    \item [(b)]training on reversed token sequences;
    \item [(c)]training on randomly (but fixed) ordered sequences. 
\end{itemize}

Without supervising the tokenizer with NTP loss, APR loss, and semantic alignment, the 1D latent space should be agnostic to ordering. However, as shown in~\cref{tab:ablate_order}, the model (a) trained on original token sequences has a much better generation FID score compared to both models that are trained with reversed and randomly ordered sequences. This shows that the end-to-end training pipeline back-propagating the NTP and APR losses learns to represent images as token sequences that are much easier to model sequentially for autoregressive models.

\vspace{10pt}
\textbf{Comparing to two-stage separate training.} We plot the next token prediction (cross entropy) loss of $\mathcal L_{\mathsf{NTP}}$ in~\cref{fig:ce_loss}. Using one-stage end-to-end training with $\mathcal L_{\mathsf{APR}}$ results in loss curves that plateau earlier and to a larger value, compared to those of separate training, while the generation FID score is much better than the latter one. This reaffirms that the NTP loss, $\mathcal L_{\mathsf{NTP}}$, cannot determine how good the generative model is, and that end-to-end training effectively optimizes for final generation quality.

\subsection{Ablation Studies}

\textbf{Sequence length.} Our 1D tokenizer offers the flexibility to compress images into tokens of varying sequence lengths. As shown in~\cref{tab:ablate_seq}, using a longer sequence length consistently improves the reconstruction quality (i.e., rFID). However, continuing to increase the sequence length makes autoregressive modeling more difficult, and the gFID score peaks at 192, setting up a reconstruction-generation trade-off similar to what was observed in latent channels of latent diffusion models~\cite{yao2025taming}.
We also find that nested dropout, introduced in~\cite{flextok,pcatok}, is effective for regularizing the token space and expanding the reconstruction-generation frontier, as shown in~\cref{appendix:ablation}.

\begin{wraptable}{r}{.52\linewidth}
    \centering
    \vspace{-10pt}
    \caption{\textbf{Ablation on the choice of codebook size.} Reducing the codebook size trades off reconstruction for generation quality. Our model retains a high code usage when using a larger codebook size.}
    \label{tab:ablate_code}
    \resizebox{\linewidth}{!}{
    \begin{tabular}{lccccc}
    \toprule
        Codebook size & 1024 & 2048 & 4096 & 8192 & 16384 \\
        \midrule
        \hspace{1.4pt}rFID $\downarrow$ & 1.18 & 1.07 & 1.02 & 0.98 & \textbf{0.96} \\
        gFID (w/o guidance) $\downarrow$ & 3.24 & \textbf{3.20} & 3.32 & 3.68 & 4.08 \\
        Code Usage $\uparrow$ & 100\% & 100\% & 99.7\% & 99.7\% & 99.2\% \\
        \bottomrule
    \end{tabular}
    }
    
\end{wraptable}
\vspace{10pt}
\textbf{Codebook size.} We study the effect of choosing different codebook sizes for latent vector quantization. We train a \ours-L model for 50 epochs with a codebook size of $K \in [1024, 2048, 4096, 8192, 16384]$. We observe yet another tradeoff between reconstruction and generation as shwon in~\cref{tab:ablate_code}. While increasing the number of codes improves the reconstruction FID, it also complicates the next token classification task for AR models.

We conduct additional experiments on training larger models with codebook sizes of 4096 and 16384. The convergence curves of gFID are shown in~\cref{fig:gfid_converge}. We find that using a larger model (e.g., \ours-H) closes the gap between the two codebook sizes as the model converges. Specifically, final gFID scores on \ours-L are $1.74$ for $K=4096$ and $1.92$ for $K=16384$, but the results on \ours-H are gFID $=1.48$ and $1.51$, respectively. This indicates that the optimal codebook size $K$ shifts as the AR model's capacity increases, and that scaling up the \ours~model can alleviate the reconstruction-generation dilemma in choosing codebook size.

\vspace{10pt}
\textbf{More experimental results.} In~\cref{appendix:ablation}, we present additional experimental results, including ImageNet 512 generation results and more reconstruction metrics; and more ablation studies, including choosing different loss weights, dropout strategies, and vision foundation models for representation alignment.

\section{Conclusion}

We present an \textbf{\textit{end-to-end}} training pipeline for autoregressive image generation, which jointly optimizes a 1D vision tokenizer and an AR generative model for \textbf{\textit{reconstruction, generation, and semantic alignment}}. We find that the next token prediction loss of the AR model on discrete latent sequences cannot determine its final generation quality, and thus propose an \textbf{\textit{autoregressive prediction reconstruction}} loss to bridge this gap. We investigate ways to incorporate global semantic information from VFMs into our 1D vision tokenizer, without enforcing the sequential latent space to align with the 2D spatial structure of VFM representations. Built on top of this training pipeline, our \ours~tokenizer learns a 1D sequential token space that facilitates autoregressive modeling, and significantly improves the overall generation quality of AR models. We instantiate \ours~across four model sizes, demonstrating a scaling property and achieving a gFID score of 1.48 without guidance. 

\clearpage

\bibliographystyle{plainnat}
\bibliography{main}

@String(ICLR = {Int. Conf. Learn. Represent.})

@String(ICLR  = {ICLR})

@inproceedings{rombach2022high,
  title={High-resolution image synthesis with latent diffusion models},
  author={Rombach, Robin and Blattmann, Andreas and Lorenz, Dominik and Esser, Patrick and Ommer, Bj{\"o}rn},
  booktitle={Proceedings of the IEEE/CVF conference on computer vision and pattern recognition},
  pages={10684--10695},
  year={2022}
}

@inproceedings{peebles2023scalable,
  title={Scalable diffusion models with transformers},
  author={Peebles, William and Xie, Saining},
  booktitle={Proceedings of the IEEE/CVF international conference on computer vision},
  pages={4195--4205},
  year={2023}
}

@article{dhariwal2021diffusion,
  title={Diffusion models beat gans on image synthesis},
  author={Dhariwal, Prafulla and Nichol, Alexander},
  journal={Advances in neural information processing systems},
  volume={34},
  pages={8780--8794},
  year={2021}
}

@misc{simeoni2025dinov3,
  title={{DINOv3}},
  author={Sim{\'e}oni, Oriane and Vo, Huy V. and Seitzer, Maximilian and Baldassarre, Federico and Oquab, Maxime and Jose, Cijo and Khalidov, Vasil and Szafraniec, Marc and Yi, Seungeun and Ramamonjisoa, Micha{\"e}l and Massa, Francisco and Haziza, Daniel and Wehrstedt, Luca and Wang, Jianyuan and Darcet, Timoth{\'e}e and Moutakanni, Th{\'e}o and Sentana, Leonel and Roberts, Claire and Vedaldi, Andrea and Tolan, Jamie and Brandt, John and Couprie, Camille and Mairal, Julien and J{\'e}gou, Herv{\'e} and Labatut, Patrick and Bojanowski, Piotr},
  year={2025},
  eprint={2508.10104},
  archivePrefix={arXiv},
  primaryClass={cs.CV},
  url={https://arxiv.org/abs/2508.10104},
}

@misc{chen2023pixartalpha,
      title={PixArt-$\alpha$: Fast Training of Diffusion Transformer for Photorealistic Text-to-Image Synthesis}, 
      author={Junsong Chen and Jincheng Yu and Chongjian Ge and Lewei Yao and Enze Xie and Yue Wu and Zhongdao Wang and James Kwok and Ping Luo and Huchuan Lu and Zhenguo Li},
      year={2023},
      eprint={2310.00426},
      archivePrefix={arXiv},
      primaryClass={cs.CV}
}

@inproceedings{yu2025repa,
  title={Representation Alignment for Generation: Training Diffusion Transformers Is Easier Than You Think},
  author={Sihyun Yu and Sangkyung Kwak and Huiwon Jang and Jongheon Jeong and Jonathan Huang and Jinwoo Shin and Saining Xie},
  year={2025},
  booktitle={ICLR},
}

@inproceedings{radford2021clip,
  title={Learning transferable visual models from natural language supervision},
  author={Radford, Alec and Kim, Jong Wook and Hallacy, Chris and Ramesh, Aditya and Goh, Gabriel and Agarwal, Sandhini and Sastry, Girish and Askell, Amanda and Mishkin, Pamela and Clark, Jack and others},
  booktitle={ICML},
  pages={8748--8763},
  year={2021},
  organization={PmLR}
}

@article{oquab2023dinov2,
  title={Dinov2: Learning robust visual features without supervision},
  author={Oquab, Maxime and Darcet, Timoth{\'e}e and Moutakanni, Th{\'e}o and Vo, Huy and Szafraniec, Marc and Khalidov, Vasil and Fernandez, Pierre and Haziza, Daniel and Massa, Francisco and El-Nouby, Alaaeldin and others},
  journal={arXiv preprint arXiv:2304.07193},
  year={2023}
}

@article{titok,
  title={An image is worth 32 tokens for reconstruction and generation},
  author={Yu, Qihang and Weber, Mark and Deng, Xueqing and Shen, Xiaohui and Cremers, Daniel and Chen, Liang-Chieh},
  journal={Advances in Neural Information Processing Systems},
  volume={37},
  pages={128940--128966},
  year={2025}
}

@misc{yao2025taming,
      title={Reconstruction vs. Generation: Taming Optimization Dilemma in Latent Diffusion Models}, 
      author={Jingfeng Yao and Xinggang Wang},
      year={2025},
      eprint={2501.01423},
      archivePrefix={arXiv},
      primaryClass={cs.CV},
      url={https://arxiv.org/abs/2501.01423}, 
}

@inproceedings{maskgit,
  title={Maskgit: Masked generative image transformer},
  author={Chang, Huiwen and Zhang, Han and Jiang, Lu and Liu, Ce and Freeman, William T},
  booktitle={Proceedings of the IEEE/CVF conference on computer vision and pattern recognition},
  pages={11315--11325},
  year={2022}
}

@article{llamagen,
  title={Autoregressive model beats diffusion: Llama for scalable image generation},
  author={Sun, Peize and Jiang, Yi and Chen, Shoufa and Zhang, Shilong and Peng, Bingyue and Luo, Ping and Yuan, Zehuan},
  journal={arXiv preprint arXiv:2406.06525},
  year={2024}
}

@article{magvitv2,
  title={Language Model Beats Diffusion--Tokenizer is Key to Visual Generation},
  author={Yu, Lijun and Lezama, Jos{\'e} and Gundavarapu, Nitesh B and Versari, Luca and Sohn, Kihyuk and Minnen, David and Cheng, Yong and Birodkar, Vighnesh and Gupta, Agrim and Gu, Xiuye and others},
  journal={arXiv preprint arXiv:2310.05737},
  year={2023}
}

@article{mar,
  title={Autoregressive image generation without vector quantization},
  author={Li, Tianhong and Tian, Yonglong and Li, He and Deng, Mingyang and He, Kaiming},
  journal={Advances in Neural Information Processing Systems},
  volume={37},
  pages={56424--56445},
  year={2024}
}

@article{maskdit,
  title={Fast training of diffusion models with masked transformers},
  author={Zheng, Hongkai and Nie, Weili and Vahdat, Arash and Anandkumar, Anima},
  journal={arXiv preprint arXiv:2306.09305},
  year={2023}
}

@article{svg,
  title={Latent Diffusion Model without Variational Autoencoder},
  author={Shi, Minglei and Wang, Haolin and Zheng, Wenzhao and Yuan, Ziyang and Wu, Xiaoshi and Wang, Xintao and Wan, Pengfei and Zhou, Jie and Lu, Jiwen},
  journal={arXiv preprint arXiv:2510.15301},
  year={2025}
}

@article{rae,
  title={Diffusion Transformers with Representation Autoencoders},
  author={Zheng, Boyang and Ma, Nanye and Tong, Shengbang and Xie, Saining},
  journal={arXiv preprint arXiv:2510.11690},
  year={2025}
}

@article{chen2025aligning,
  title={Aligning Visual Foundation Encoders to Tokenizers for Diffusion Models},
  author={Chen, Bowei and Bi, Sai and Tan, Hao and Zhang, He and Zhang, Tianyuan and Li, Zhengqi and Xiong, Yuanjun and Zhang, Jianming and Zhang, Kai},
  journal={arXiv preprint arXiv:2509.25162},
  year={2025}
}

@article{bi2025vision,
  title={Vision Foundation Models Can Be Good Tokenizers for Latent Diffusion Models},
  author={Bi, Tianci and Zhang, Xiaoyi and Lu, Yan and Zheng, Nanning},
  journal={arXiv preprint arXiv:2510.18457},
  year={2025}
}

@inproceedings{rqvae,
  title={Autoregressive image generation using residual quantization},
  author={Lee, Doyup and Kim, Chiheon and Kim, Saehoon and Cho, Minsu and Han, Wook-Shin},
  booktitle={Proceedings of the IEEE/CVF Conference on Computer Vision and Pattern Recognition},
  pages={11523--11532},
  year={2022}
}

@article{vitvqgan,
  title={Vector-quantized image modeling with improved vqgan},
  author={Yu, Jiahui and Li, Xin and Koh, Jing Yu and Zhang, Han and Pang, Ruoming and Qin, James and Ku, Alexander and Xu, Yuanzhong and Baldridge, Jason and Wu, Yonghui},
  journal={arXiv preprint arXiv:2110.04627},
  year={2021}
}

@inproceedings{deng2009imagenet,
  title={Imagenet: A large-scale hierarchical image database},
  author={Deng, Jia and Dong, Wei and Socher, Richard and Li, Li-Jia and Li, Kai and Fei-Fei, Li},
  booktitle={2009 IEEE conference on computer vision and pattern recognition},
  pages={248--255},
  year={2009},
  organization={Ieee}
}

@article{kingma2014adam,
  title={Adam: A method for stochastic optimization},
  author={Kingma, Diederik P and Ba, Jimmy},
  journal={arXiv preprint arXiv:1412.6980},
  year={2014}
}

@article{heusel2017gans,
  title={Gans trained by a two time-scale update rule converge to a local nash equilibrium},
  author={Heusel, Martin and Ramsauer, Hubert and Unterthiner, Thomas and Nessler, Bernhard and Hochreiter, Sepp},
  journal={Advances in neural information processing systems},
  volume={30},
  year={2017}
}

@article{salimans2016improved,
  title={Improved techniques for training gans},
  author={Salimans, Tim and Goodfellow, Ian and Zaremba, Wojciech and Cheung, Vicki and Radford, Alec and Chen, Xi},
  journal={Advances in neural information processing systems},
  volume={29},
  year={2016}
}

@misc{vae,
  title={Auto-encoding variational bayes},
  author={Kingma, Diederik P and Welling, Max and others},
  year={2013},
  publisher={Banff, Canada}
}

@inproceedings{lpips,
  title={The unreasonable effectiveness of deep features as a perceptual metric},
  author={Zhang, Richard and Isola, Phillip and Efros, Alexei A and Shechtman, Eli and Wang, Oliver},
  booktitle={Proceedings of the IEEE conference on computer vision and pattern recognition},
  pages={586--595},
  year={2018}
}

@article{leng2025repae,
  title={Repa-e: Unlocking vae for end-to-end tuning with latent diffusion transformers},
  author={Leng, Xingjian and Singh, Jaskirat and Hou, Yunzhong and Xing, Zhenchang and Xie, Saining and Zheng, Liang},
  journal={arXiv preprint arXiv:2504.10483},
  year={2025}
}

@article{van2017neural,
  title={Neural discrete representation learning},
  author={Van Den Oord, Aaron and Vinyals, Oriol and others},
  journal={Advances in neural information processing systems},
  volume={30},
  year={2017}
}

@inproceedings{esser2021taming,
  title={Taming transformers for high-resolution image synthesis},
  author={Esser, Patrick and Rombach, Robin and Ommer, Bjorn},
  booktitle={Proceedings of the IEEE/CVF conference on computer vision and pattern recognition},
  pages={12873--12883},
  year={2021}
}

@article{razavi2019generating,
  title={Generating diverse high-fidelity images with vq-vae-2},
  author={Razavi, Ali and Van den Oord, Aaron and Vinyals, Oriol},
  journal={Advances in neural information processing systems},
  volume={32},
  year={2019}
}

@inproceedings{
vit,
title={An Image is Worth 16x16 Words: Transformers for Image Recognition at Scale},
author={Alexey Dosovitskiy and Lucas Beyer and Alexander Kolesnikov and Dirk Weissenborn and Xiaohua Zhai and Thomas Unterthiner and Mostafa Dehghani and Matthias Minderer and Georg Heigold and Sylvain Gelly and Jakob Uszkoreit and Neil Houlsby},
booktitle={International Conference on Learning Representations},
year={2021},
url={https://openreview.net/forum?id=YicbFdNTTy}
}

@article{ge2023making,
  title={Making llama see and draw with seed tokenizer},
  author={Ge, Yuying and Zhao, Sijie and Zeng, Ziyun and Ge, Yixiao and Li, Chen and Wang, Xintao and Shan, Ying},
  journal={arXiv preprint arXiv:2310.01218},
  year={2023}
}

@inproceedings{flextok,
  title={FlexTok: Resampling Images into 1D Token Sequences of Flexible Length},
  author={Bachmann, Roman and Allardice, Jesse and Mizrahi, David and Fini, Enrico and Kar, O{\u{g}}uzhan Fatih and Amirloo, Elmira and El-Nouby, Alaaeldin and Zamir, Amir and Dehghan, Afshin},
  booktitle={Forty-second International Conference on Machine Learning},
  year={2025}
}

@article{pcatok,
  title={" Principal Components" Enable A New Language of Images},
  author={Wen, Xin and Zhao, Bingchen and Elezi, Ismail and Deng, Jiankang and Qi, Xiaojuan},
  journal={arXiv preprint arXiv:2503.08685},
  year={2025}
}

@inproceedings{IBQ,
  title={Scalable image tokenization with index backpropagation quantization},
  author={Shi, Fengyuan and Luo, Zhuoyan and Ge, Yixiao and Yang, Yujiu and Shan, Ying and Wang, Limin},
  booktitle={Proceedings of the IEEE/CVF International Conference on Computer Vision},
  pages={16037--16046},
  year={2025}
}

@inproceedings{dqvae,
  title={Towards accurate image coding: Improved autoregressive image generation with dynamic vector quantization},
  author={Huang, Mengqi and Mao, Zhendong and Chen, Zhuowei and Zhang, Yongdong},
  booktitle={Proceedings of the IEEE/CVF Conference on Computer Vision and Pattern Recognition},
  pages={22596--22605},
  year={2023}
}

@article{FSQ,
  title={Finite scalar quantization: Vq-vae made simple},
  author={Mentzer, Fabian and Minnen, David and Agustsson, Eirikur and Tschannen, Michael},
  journal={arXiv preprint arXiv:2309.15505},
  year={2023}
}

@article{zhu2024scaling,
  title={Scaling the codebook size of VQ-GAN to 100,000 with a utilization rate of 99\%},
  author={Zhu, Lei and Wei, Fangyun and Lu, Yanye and Chen, Dong},
  journal={Advances in Neural Information Processing Systems},
  volume={37},
  pages={12612--12635},
  year={2024}
}

@article{VAR,
  title={Visual autoregressive modeling: Scalable image generation via next-scale prediction},
  author={Tian, Keyu and Jiang, Yi and Yuan, Zehuan and Peng, Bingyue and Wang, Liwei},
  journal={Advances in neural information processing systems},
  volume={37},
  pages={84839--84865},
  year={2024}
}

@article{gigatok,
  title={Gigatok: Scaling visual tokenizers to 3 billion parameters for autoregressive image generation},
  author={Xiong, Tianwei and Liew, Jun Hao and Huang, Zilong and Feng, Jiashi and Liu, Xihui},
  journal={arXiv preprint arXiv:2504.08736},
  year={2025}
}

@article{gpt3,
  title={Language models are few-shot learners},
  author={Brown, Tom and Mann, Benjamin and Ryder, Nick and Subbiah, Melanie and Kaplan, Jared D and Dhariwal, Prafulla and Neelakantan, Arvind and Shyam, Pranav and Sastry, Girish and Askell, Amanda and others},
  journal={Advances in neural information processing systems},
  volume={33},
  pages={1877--1901},
  year={2020}
}

@article{karras2024guiding,
  title={Guiding a diffusion model with a bad version of itself},
  author={Karras, Tero and Aittala, Miika and Kynk{\"a}{\"a}nniemi, Tuomas and Lehtinen, Jaakko and Aila, Timo and Laine, Samuli},
  journal={Advances in Neural Information Processing Systems},
  volume={37},
  pages={52996--53021},
  year={2024}
}

@article{ho2022classifier,
  title={Classifier-free diffusion guidance},
  author={Ho, Jonathan and Salimans, Tim},
  journal={arXiv preprint arXiv:2207.12598},
  year={2022}
}

@inproceedings{sauer2023stylegan,
  title={Stylegan-t: Unlocking the power of gans for fast large-scale text-to-image synthesis},
  author={Sauer, Axel and Karras, Tero and Laine, Samuli and Geiger, Andreas and Aila, Timo},
  booktitle={International conference on machine learning},
  pages={30105--30118},
  year={2023},
  organization={PMLR}
}

@inproceedings{tseng2021regularizing,
  title={Regularizing generative adversarial networks under limited data},
  author={Tseng, Hung-Yu and Jiang, Lu and Liu, Ce and Yang, Ming-Hsuan and Yang, Weilong},
  booktitle={Proceedings of the IEEE/CVF conference on computer vision and pattern recognition},
  pages={7921--7931},
  year={2021}
}

@article{gan,
  title={Generative adversarial nets},
  author={Goodfellow, Ian J and Pouget-Abadie, Jean and Mirza, Mehdi and Xu, Bing and Warde-Farley, David and Ozair, Sherjil and Courville, Aaron and Bengio, Yoshua},
  journal={Advances in neural information processing systems},
  volume={27},
  year={2014}
}

@inproceedings{yu2025randomized,
  title={Randomized autoregressive visual generation},
  author={Yu, Qihang and He, Ju and Deng, Xueqing and Shen, Xiaohui and Chen, Liang-Chieh},
  booktitle={Proceedings of the IEEE/CVF International Conference on Computer Vision},
  pages={18431--18441},
  year={2025}
}

@article{wu2025alitok,
  title={AliTok: Towards Sequence Modeling Alignment between Tokenizer and Autoregressive Model},
  author={Wu, Pingyu and Zhu, Kai and Liu, Yu and Tang, Longxiang and Yang, Jian and Peng, Yansong and Zhai, Wei and Cao, Yang and Zha, Zheng-Jun},
  journal={arXiv preprint arXiv:2506.05289},
  year={2025}
}

@article{huang2025spectralar,
  title={SpectralAR: Spectral Autoregressive Visual Generation},
  author={Huang, Yuanhui and Chen, Weiliang and Zheng, Wenzhao and Duan, Yueqi and Zhou, Jie and Lu, Jiwen},
  journal={arXiv preprint arXiv:2506.10962},
  year={2025}
}

@article{zheng2025vision,
  title={Vision foundation models as effective visual tokenizers for autoregressive image generation},
  author={Zheng, Anlin and Wen, Xin and Zhang, Xuanyang and Ma, Chuofan and Wang, Tiancai and Yu, Gang and Zhang, Xiangyu and Qi, Xiaojuan},
  journal={arXiv preprint arXiv:2507.08441},
  year={2025}
}

@article{zhang2026restok,
  title={ResTok: Learning Hierarchical Residuals in 1D Visual Tokenizers for Autoregressive Image Generation},
  author={Zhang, Xu and Da, Cheng and Yang, Huan and Gai, Kun and Lu, Ming and Ma, Zhan},
  journal={arXiv preprint arXiv:2601.03955},
  year={2026}
}

@article{zhang2019root,
  title={Root mean square layer normalization},
  author={Zhang, Biao and Sennrich, Rico},
  journal={Advances in neural information processing systems},
  volume={32},
  year={2019}
}

@article{shazeer2020glu,
  title={Glu variants improve transformer},
  author={Shazeer, Noam},
  journal={arXiv preprint arXiv:2002.05202},
  year={2020}
}

@article{simonyan2014very,
  title={Very deep convolutional networks for large-scale image recognition},
  author={Simonyan, Karen and Zisserman, Andrew},
  journal={arXiv preprint arXiv:1409.1556},
  year={2014}
}

@article{tschannen2025siglip,
  title={SigLIP 2: Multilingual Vision-Language Encoders with Improved Semantic Understanding, Localization, and Dense Features},
  author={Tschannen, Michael and Gritsenko, Alexey and Wang, Xiao and Naeem, Muhammad Ferjad and Alabdulmohsin, Ibrahim and Parthasarathy, Nikhil and Evans, Talfan and Beyer, Lucas and Xia, Ye and Mustafa, Basil and H\'enaff, Olivier and Harmsen, Jeremiah and Steiner, Andreas and Zhai, Xiaohua},
  year={2025},
  journal={arXiv preprint arXiv:2502.14786}
}

\clearpage

\beginappendix
\section{Implementation Details}

\subsection{Model Architecture}
\label{appendix:model_config}

\subsubsection{Tokenizer.} Our 1D ViT tokenizer uses a similar architecture as TiTok~\cite{titok} to compress images into 1D discrete sequences without 2D structural prior.

\vspace{10pt}
\textbf{Encoder.} Images are first patchified into tokens with a $P \times P$ patch size and hidden dimension of $D$, resulting in a total of $HW/P^2$ patch tokens in a 2D grid. These tokens are flattened and concatenated with $L$ learnable query tokens $\rvq\in \mathbb R^{L\times D}$. We use learnable positional embeddings for simplicity.
The embedded tokens of length $HW/P^2 + L$ are then passed into multiple layers of transformer blocks. It outputs $[\rvh_{\text{Enc}}, \rvz]$ where $\rvh_{\text{Enc}} \in \mathbb R^{HW/P^2 \times D}$ is the hidden 2D embedding. The query tokens go through a linear projection layer and are mapped to $d$-dimension, i.e., $\rvz\in \mathbb R^{L \times d}$.

\vspace{10pt}
\textbf{Decoder.} The decoder follows a symmetric design of the encoder. Sequential 1D latent codes are concatenated with $HW/P^2$ mask tokens in a 2D grid. After multiple layers of transformer blocks, the decoder outputs the unmasked tokens, followed by an unpatchify layer and a convolutional output layer.

\vspace{10pt}
\textbf{Hybrid attention mask.} On our 1D ViT encoder, the attention is bidirectional among 2D patch tokens and causal along the 1D query tokens. Query tokens are allowed to attend to patch tokens, whereas patch tokens cannot attend to query tokens. Similarly, on the decoder, the attention is causal along 1D query tokens and bidirectional among 2D patch tokens. 

\vspace{10pt}
\textbf{Quantizer.} We use IBQ~\cite{IBQ} to quantize latent embeddings. We stabilize the codebook training by applying an $\ell_2$ normalization on both the codebook $\mathcal C$ and the continuous latent embedding when computing their similarity, i.e.,

\begin{equation}
    \mathsf{logits} = \left[\frac{\rvz^T \mathcal C_1}{\|\rvz\|_2\|\mathcal C_1\|_2}, \ldots,\frac{\rvz^T \mathcal C_1}{\|\rvz\|_2\|\mathcal C_K\|_2} \right], \ \text{and}\ \rvp = \mathsf{softmax}(\mathsf{logits}/\tau).
\end{equation}

The detailed model configurations for each model size (S, B, L, H) are listed in~\cref{tab:model_config}.

\subsubsection{Autoregressive Generative Model.} 
We slightly modify the language model in LlamaGen~\cite{llamagen} for autoregressive generation. Specifically, we add an additional shared global AdaLN~\cite{peebles2023scalable} modulation with per-block learnable biases. The model applies RMSNorm~\cite{zhang2019root} and SwiGLU activation function~\cite{shazeer2020glu}. Since we tokenize images into 1D sequences, we replace the 2D RoPE with learnable positional embeddings. We include more details on model configurations in~\cref{tab:model_config}.

\subsection{Training Details}

\subsubsection{Loss functions}

Our end-to-end framework jointly trains a tokenizer with encoder $\mathcal E_\vphi$ and decoder $\mathcal D_{\vpsi}$, and an AR model $\mathcal G_\vtheta$. The training losses include reconstruction, next-token prediction (NTP), autoregressive prediction reconstruction (APR), and semantic alignment. Putting it together, we summarize the objective function of \ours~as: 

\begin{align}
    \mathcal L_{\mathrm{EOSTok}}(\vphi, \vpsi, \vtheta) &= \mathcal L_{\mathsf{VQVAE}} (\vphi, \vpsi)+ \lambda_{\mathsf{NTP}} \mathcal L_{\mathsf{NTP}}(\vphi,\vtheta) + \lambda_{\mathsf{APR}} \mathcal L_{\mathsf{APR}} (\vphi, \vpsi, \vtheta) \nonumber\\
    &+ \min_{\vomega_1,\vomega_2} \lambda_{\mathsf{sem}} (\mathcal L_{\mathsf{implicit}}(\vomega_1 ,\vphi) + \mathcal L_{\mathsf{decoder} -\mathsf{align}}(\vomega_2,\vphi,\vpsi)).
\end{align}
$\lambda_{\mathsf{NTP}}, \lambda_{\mathsf{APR}}$ and $\lambda_{\mathsf{sem}}$ are pre-determined loss weights, and $\vomega_1,\vomega_2$ are learnable MLP layers for representation alignment.
The VAE loss can be decomposed into $\mathcal L_{\mathsf{VQVAE}} = L_2 + \mathrm{LPIPS} + \lambda_{\mathsf{GAN}}\mathcal L_\mathsf{GAN} +  \lambda_{\mathsf{reg}} \mathcal L_{\mathsf{reg}}$, and $\mathcal L_{\mathsf{APR}} = L_2 + \mathrm{LPIPS}$. Furthermore, we use the discriminator from StyleGAN-T~\cite{sauer2023stylegan} for GAN loss, with LeCAM divergence~\cite{tseng2021regularizing} to stabilize training. The perceptual LPIPS loss is computed with a VGG~\cite{simonyan2014very} backbone. We include details of loss weights in~\cref{tab:model_config}.

\subsubsection{Optimizer}
We use Adam optimizers with an initial learning rate of 1e-4 for both the tokenizer and the AR model, and employ a cosine learning rate scheduler that decays to 1e-6 at 2M iterations. For the discriminator, we use an Adam optimizer with a fixed learning rate of 1e-4.
We train our models on 8 H100 GPUs with a batch size of 256 for 400 epochs (approximately 2M iterations). More training details can be found in~\cref{tab:model_config}.

\subsection{Sampling Details}

We apply autoregressive sampling with KV cache, using a temperature of 1.0 without top-k or top-p strategies. 
We use classifier guidance on \ours-S and \ours-B, i.e., $\ell_g = \ell_u + s ( \ell_c - \ell_u)$, where $\ell_u$ and $\ell_c$ are logits of unconditional and class conditional sampling, and $s$ is the CFG scale.
We notice a diminishing effect on applying CFG sampling when scaling up our model to \ours-L and \ours-H as the unconditional generation quality of the model improves. We thus apply AutoGuidance following~\cite{karras2024guiding} by training a lightweight AR model on the tokenizer and using it to replace the unconditional logits. We use the same budget for all models to search for the best guidance scale. 

\subsection{Computation Cost}

We provide a detailed cost analysis of EOSTok in~\cref{tab:train-cost}, including the GFLOPs of each module and the peak memory usage in training. Moreover, the sampling GFLOPs of our largest model EOSTok-H are listed in~\cref{tab:eval-cost}. Since we use KV cache during sampling, the sampling speed is much faster than diffusion sampling. With batched image generation, EOSTok-H can generate about 10.5 images per second on a single H100. As shown in~\cref{tab:eval-cost}, EOSTok-H is 20 to 100 times faster than the DiT-XL/2 model, depending on the diffusion sampling algorithm used.

\begin{table}[t]
    \centering
    \caption{\textbf{Computation and memory cost.}}
    \label{tab:train-cost}
    \resizebox{0.6\linewidth}{!}{
    \begin{tabular}{lcc}
    \toprule
    & EOSTok-L & EOSTok-H \\
    \midrule
       Tokenizer GFLOPs (Encoder + Decoder) &  91 + 91 & 210 + 210\\
       AR Transformer GFLOPs &  162 & 338 \\
       VFM embedder GFLOPs & 162 & 162 \\
       Forward total (AR + Encode + $2\times$ Decode + VFM) & 597 & 1130 \\
       \midrule
       Overhead compared to two-stage training & 15.2\% & 18.6\% \\
       Peak memory (Per device batch size 32) & 36.51 GB & 56.96 GB \\
    \bottomrule
    \end{tabular}}
\end{table}

\begin{table}[t]
    \centering
    \caption{\textbf{Sampling efficiency of EOSTok.} (*The original paper uses a different convention of 1 mul-add $=$ 1 op, which halves the number.)}
    \label{tab:eval-cost}
    \resizebox{0.5\linewidth}{!}{
    \begin{tabular}{lccc}
    \toprule
    EOSTok-H & AR Model (256 tokens) & Decoder & Total \\
    \midrule
    GFLOPs & 342.3 & 210.1 & 552.4 \\
    \midrule
    \midrule
    DiT-XL/2 & Diffusion Model & Decoder & Total \\
    \midrule
    GFLOPs* & 237.2 $\times$ 250 steps & 622.2 & 59.9k \\
    \bottomrule
    \end{tabular}}
\end{table}

\definecolor{rowhl}{HTML}{E6E6FA}

\begin{table}[t]
    \centering
    \caption{\textbf{Implementation details of \ours~on ImageNet 256 generation experiments.}}
    \label{tab:model_config}
    {\small
    \begin{tabular}{lcccc}
    \toprule
    & \ours-S & \ours-B & \ours-L & \ours-H\\
    \midrule
    \multicolumn{5}{l}{\textit{\textbf{Tokenizer configs}}} \\
    Params. & 165M &  165M &  165M & 388M\\
    Patch size & 16 & 16 & 16 & 16 \\
    Transformer layers (encoder-decoder) & 12-12 & 12-12 & 12-12 & 16-16 \\
    Hidden dimensions & 768 & 768 & 768 & 1024\\
    Attention heads & 12 & 12 & 12 & 16\\
    Latent space dimensions & 64 & 64 & 64 & 64\\
    \multicolumn{5}{l}{\textbf{\textit{Quantizer}}}\\
    Codebook size & 4096 & 4096 & 4096 &  4096\\
    Temperature & 1.0 & 1.0 & 1.0 & 1.0 \\
    Regularization weights $\lambda_{\mathsf{reg}}$ & 1e-3 & 1e-3 & 1e-3 & 1e-3 \\
    Entropy weights & 0.01 & 0.01 & 0.01 & 0.01 \\
    \midrule
    \multicolumn{5}{l}{\textbf{\textit{{Autoregressive Model configs}}}}\\
    Params. & 93M & 164M & 312M & 644M \\
    Layers & 12 & 12 & 24 & 32 \\
    Hidden dimensions & 768 & 1024 & 1024 & 1280 \\
    Attention heads & 12 & 16 & 16 & 20\\
    \midrule 
    \textbf{\textit{Training loss weights}}  \\
    Reconstruction L2 & 1.0 & 1.0 & 1.0 & 1.0 \\
    Reconstruction LPIPS & 1.0 & 1.0 & 1.0 & 1.0 \\
    Implicit Alignment & 1.0 & 1.0 & 1.0 & 1.0 \\
    GAN loss & 0.1 & 0.1 & 0.1 & 0.1 \\
    GAN LeCam & 0.05 & 0.05 & 0.05 & 0.05 \\
    APR L2 & 1.0 & 1.0 & 1.0 & 1.0 \\
    APR LPIPS & 1.0 & 1.0 & 1.0 & 1.0 \\
    Next token prediction & 0.1 & 0.1 & 0.1 & 0.01 \\
    \midrule 
    \textbf{\textit{Training details}} \\
    Batch size & 256 & 256 & 256 &  256\\
    Epochs & 400 & 400 & 400 & 400 \\
    Learning rate & 1e-4 & 1e-4 & 1e-4 & 1e-4\\
    Cosine learning rate min & 1e-6 & 1e-6 & 1e-6 & 1e-6\\ 
    Adam $\beta_1$ & 0.9 & 0.9 & 0.9 & 0.9 \\
    Adam $\beta_2$ Tokenizer/AR & 0.999/0.95 & 0.999/0.95 & 0.999/0.95 & 0.999/0.95 \\
    EMA decay rate & 0.9999 & 0.9999 & 0.9999 & 0.9999 \\ 
    Nested dropout ratio & 0.5 & 0.5 & 0.5 & 1.0 \\
    Class dropout ratio & 0.1 & 0.1 & 0.1 & 0.1\\
    
    \bottomrule
    \end{tabular}
    }

\end{table}

\clearpage
\newpage
\section{Additional Experimental Results}
\label{appendix:ablation}

\textbf{APR weights.} We further study the effect of using different APR weights in end-to-end training. We fix the NTP loss weight $\lambda_{\mathsf{NTP}}$ to 0.1 in this experiment, and train an \ours-L model for 50 epochs. As shown in~\cref{tab:ablate_apr}, using an APR weight of 1.0 achieves the best performance for both reconstruction and generation quality. 

\vspace{10pt}
\textbf{Nested dropout.} We conduct experiments to understand the effect of applying nested dropout proposed in~\cite{flextok,pcatok} to regularize the 1D sequential latent space and compress important information into tokens in the front. We consider $p = [0.0, 0.25, 0.5, 1.0]$ for the probability of applying nested dropout during training. As shown in~\cref{tab:ablate_drop}, using a nested dropout strategy can further make latent tokens easier for AR modeling. Specifically, applying nested dropout with probability $p=1$ results in a $17.6\%$ AR prediction accuracy, while it slightly worsens generation quality since the over-compressing hurts the model's ability of reconstructing images. We thus apply nested dropout with rate $p=0.5$ in our training pipeline.

\begin{table}[t]
    \centering
    \begin{minipage}[t]{0.48\linewidth}
        \centering
        \caption{\textbf{Ablation on the APR loss weight.}}
        \label{tab:ablate_apr}
        \resizebox{0.95\textwidth}{!}{
        \begin{tabular}{lccccc}
            \toprule
            APR loss weight $\lambda_{\mathsf{APR}}$ & 0.0 & 0.5 & 1.0 & 2.0 & 4.0  \\
            \midrule
            rFID $\downarrow$ & 1.03 & \textbf{1.02 }& \textbf{1.02} & 1.05 & 1.12  \\
            gFID (w/o guidance) $\downarrow$ & 4.09 & 3.52 & \textbf{3.32} & 3.34 & 3.57 \\
            \bottomrule
        \end{tabular}
        }
    \end{minipage}\hfill
    \begin{minipage}[t]{0.48\linewidth}
        \centering
        \caption{\textbf{Ablation on the nested dropout rate.}}
        \label{tab:ablate_drop}
        \resizebox{0.9\textwidth}{!}{
        \begin{tabular}{lcccc}
            \toprule
            Dropout probability & 0.0 & 0.25 & 0.5 & 1.0  \\
            \midrule
            rFID $\downarrow$ & \textbf{0.85} & 0.94 & 1.02 & 1.24  \\
            gFID (w/o guidance) $\downarrow$ & 3.70 & 3.52 & \textbf{3.32} & 3.50 \\
            AR Acc. $\uparrow$ & 10.2 \% & 10.8\% & 11.9\% & \textbf{17.6\%}  \\
            \bottomrule
        \end{tabular}
        }
    \end{minipage}
\end{table}

\vspace{10pt}
\textbf{The choice of VFM in representation alignment.} We further experiment with EOSTok-L in our ablation setting using a SigLIP2~\cite{tschannen2025siglip} model, which contains richer global semantic information. As shown in~\cref{tab:ablate-vfm}, SigLIP2 slightly improves the generative results compared to DINOv2, which shows the robustness of our framework to the choice of pretrained vision foundation models.
\begin{table}[t]
    \centering
    \begin{minipage}[t]{0.48\linewidth}
    \caption{\textbf{Ablation on the choice of vision foundation models for representation alignment.}}
    \label{tab:ablate-vfm}
\centering
\resizebox{0.85\textwidth}{!}{
    \begin{tabular}{lcc}
    \toprule
    Pretrained VFM & DINOV2 & SigLIP2 \\
    \midrule
    rFID & 1.02 & 0.88 \\
    gFID (w/o guidance)  & 3.32 & 3.02 \\
    \bottomrule
    \end{tabular}}

\vspace{1em}  

    \caption{\textbf{Scalability of EOSTok to ImageNet 512.}}
    \label{tab:result-512}
    \centering
    \resizebox{0.75\textwidth}{!}{
    \begin{tabular}{lc}
        \toprule
        Models & gFID (w/o guidance)\\
        \midrule
        DiT-XL/2~\cite{peebles2023scalable} & 12.03 \\
        MaskDiT~\cite{maskdit} & 10.79 \\
        TiTok-B-128~\cite{titok}	& 4.17 \\
        TiTok-L-64~\cite{titok} & 3.99\\
        \textbf{EOSTok-L (Ours)} &  \textbf{1.98} \\
        \bottomrule
    \end{tabular}}
    
 \end{minipage}\hfill
 \begin{minipage}[t]{0.48\linewidth}
        \centering
        \caption{\textbf{More reconstruction metrics of the EOSTok tokenizer.}}
        \label{tab:recon-metrics}
        \resizebox{\textwidth}{!}{
        \begin{tabular}{lcccc}
            \toprule
            Models & PSNR & SSIM & LPIPS & rFID \\
            \midrule
            EOSTok-L (n=4096) & 22.15 & 0.67 & 0.231 & 0.73 \\
            IBQ (n=16384) & 22.01 & 0.61 & 0.224 & 1.37 \\
            IBQ (n=262144) & 22.69 & 0.64 & 0.203 & 1.00 \\
            GigaTok-B-L (n=16384) & 21.21 & 0.68 & 0.206 & 0.81 \\
            LlamaGen (n=16384) & 20.79 & 0.62 & - & 2.19 \\
            Semanticist (continuous 1D) & 21.61 & 0.63 & - & 0.78 \\
        \bottomrule
        \end{tabular}
        }
    \end{minipage}
\end{table}



\vspace{10pt}
\textbf{Scalability to higher resolution.} We show that EOSTok can be naturally extended to higher resolution. We uses the same architecture of EOSTok-L, keeping the patch size of 16 and the AR sequence length of 256.
As shown in~\cref{tab:result-512}, our EOSTok-L achieves an gFID of 1.98 without guidance, outperforming existing diffusion models with 2D tokenizers, and also existing mask based generative models with 1D tokenizers.

\vspace{10pt}
\textbf{Reconstruction evaluations.} We include more comprehensive reconstruction metrics in~\cref{tab:recon-metrics}. We include the results of recent discrete tokenizers for reference. Our tokenizer achieves comparable performance on PSNR, SSIM, and LPIPS despite using compact 1D compression and better distributional performance, as measured by rFID.

\end{document}